\documentclass[dvipsnames,format=sigconf]{acmart}
\makeatletter
\newcommand\anonacm[1]{%
\if@ACM@anonymous
    \textcolor{red}{ANONYMIZED}%
  \else
    #1%
  \fi
}
\makeatother 

\usepackage{subfig}
\usepackage[noend]{algorithmic}
\usepackage{algorithm}
\usepackage{fancyvrb}
\usepackage{xfrac}
\usepackage{tikz}
\usetikzlibrary{positioning}
\usepackage[export]{adjustbox}
\usepackage[normalem]{ulem} 
\usepackage{array}   
\newcolumntype{R}{>{$}r<{$}} 

\newcommand{\egp}{eggp}

\AtBeginDocument{%
  \providecommand\BibTeX{{%
    \normalfont B\kern-0.5em{\scshape i\kern-0.25em b}\kern-0.8em\TeX}}}

\copyrightyear{2025}
\acmYear{2025}
\setcopyright{acmlicensed}\acmConference[GECCO '25]{Genetic and Evolutionary Computation Conference}{July 14--18, 2025}{Malaga, Spain}
\acmBooktitle{Genetic and Evolutionary Computation Conference (GECCO '25), July 14--18, 2025, Malaga, Spain}
\acmDOI{10.1145/3712256.3726383}
\acmISBN{979-8-4007-1465-8/2025/07}

\begin{document}

\title{Improving Genetic Programming for Symbolic Regression with Equality Graphs}

\author{Fabrício Olivetti de França}
\orcid{0000-0002-2741-8736}
\affiliation{%
  \institution{Federal University of ABC}
  \city{Santo Andre}
  \state{São Paulo}
  \country{Brazil}
}
\email{folivetti@ufabc.edu.br}

\author{Gabriel Kronberger}
\orcid{0000-0002-3012-3189}
\affiliation{%
  \institution{University of Applied Sciences Upper Austria}
  \city{Hagenberg}
  \state{Upper Austria}
  \country{Austria}
}
\email{gabriel.kronberger@fh-hagenberg.at}


\begin{abstract} 
The search for symbolic regression models with genetic programming (GP) has a tendency of revisiting expressions in their original or equivalent forms. Repeatedly evaluating equivalent expressions is inefficient, as it does not immediately lead to better solutions.
However, evolutionary algorithms require diversity and should allow the accumulation of inactive building blocks that can play an important role at a later point. 
The equality graph is a data structure capable of compactly storing expressions and their equivalent forms allowing an efficient verification of whether an expression has been visited in any of their stored equivalent forms.
We exploit the e-graph to adapt the subtree operators to reduce the chances of revisiting expressions. Our adaptation, called \egp{}, stores every visited expression in the e-graph, allowing us to filter out from the available selection of subtrees all the combinations that would create already visited expressions. 
Results show that, for small expressions, this approach improves the performance of a simple GP algorithm to compete with PySR and Operon without increasing computational cost. As a highlight, \egp{} was capable of reliably delivering short and at the same time accurate models for a selected set of benchmarks from SRBench and a set of real-world datasets.
\end{abstract}




\maketitle

\section{Introduction}
Symbolic regression (SR)~\cite{Koza1992,kronberger2024} searches for a mathematical function that approximates a set of data points 
often used for scientific discovery~\cite{schmidt2009distilling,kronberger2024,cranmerpysr, udrescu2020ai,russeil2024multiview,caolearning}.
The current SotA~\cite{la2021contemporary,de2024srbench} uses genetic programming (GP) as the main search engine incorporating numerical parameters that can be fitted into the data using optimization techniques.
The search for a symbolic model is NP-hard~\cite{virgolinsymbolic} and when searching for a parametric model, it also requires the solution to a multimodal optimization problem, which by itself is NP-hard~\cite{murty1985some} and can hinder the search for the optimal solution. 

To make matters worse, the usual way of encoding mathematical expressions as symbolic expression trees, allows GP to visit semantically equivalent expressions\footnote{in this paper, we will refer to \emph{semantically equivalence} as simply \emph{equivalence}.} with different syntax~\cite{kronberger2024inefficiency}. These equivalent expressions may be unnecessarily large and with redundant parameters, reducing the probability of finding their optimal values~\cite{de2023reducing,kronberger2024jsc}. Even for the simple expression $p_1 x_1$ we can produce an infinite number of equivalent expressions considering that $p$ are fitting parameters, for example $((p_1 x_1) + (p_2 x_1)$, $x_1 / p_1$, $x_1^2 / (p_1 x_1)$, are all different parameterizations of the same expression.

GP cannot easily differentiate between equivalent expressions, and applying simplification heuristics are often insufficient, as seen in~\cite{de2023reducing}.
Some authors~\cite{keller1999evolution,milleratall2006} argue that redundancy is necessary to allow the algorithm to navigate through the search space, as these equivalent expressions are guaranteed to have the same accuracy, allowing the search to keep multiple genetically different variations of solution candidates in the hopes of finding better solutions. However, this supposition could not be validate as there were no means to efficiently verify equivalence.

Equality saturation~\cite{willsey2021egg} can produce all the equivalent expressions of a given expression through the parallel application of a set of equivalence rules. Given an expression represented as a directed acyclic graph, and a set of equivalence rules, it  iteratively applies the rules and stores all equivalent programs in a compact data structure called \emph{equality graph (e-graph)}. The main idea is that upon saturation, the graph will contain all equivalent forms of the original program and the optimal form can be extracted from the e-graph using a heuristic cost function.
This technique was previously used in the context of SR in~\cite{de2023reducing,kronberger2024jsc} to investigate the problem of overparameterization that can negatively affect the fitting of numerical parameters. The e-graph has another feature that can be exploited by SR algorithms: it implements an efficient pattern matching algorithm that can answer whether a given expression, or any of its equivalents, are already stored in the e-graph structure.

In this work we introduce \egp{}, a GP algorithm that exploits the pattern matching capabilities of the e-graph to try to enforce the generation of \emph{unvisited expressions} when applying the crossover and mutation operators. In this context, unvisited expressions mean any expression, or their equivalents, that was not previously evaluated during the history of search. In short, after choosing the crossover point of the first parent, the choices of points of the second parent are limited to those that will ensure the generation of an unvisited expression. For the mutation, after choosing a node of the expression at random, it will generate a new subtree, limiting the choice of its root node to the set that will ensure the generation of an unvisited expression. This procedure will enforce the introduction of novel solutions not only w.r.t. the current population but to the entire history of the search.


The research questions we want to address in this paper are:

\begin{enumerate}
    \item What is the impact of increasing the probability of generating novelty, when compared to a minimalist implementation of GP for SR?
    \item How close does \egp{} get to the state-of-the-art without resorting to more advanced concepts such as specialized mutation operators, enforcing the placement of numerical parameters, and promoting diversity throug island model?
\end{enumerate}

These operators are tested inside a minimalist GP implementation, and compared against this same algorithm with the original subtree operators, and two high performant algorithms: Operon~\cite{burlacu2020operon} and PySR~\cite{cranmerpysr}. The results show that this \emph{simple} modification, provided we have a working implementation of the e-graph, can improve the performance of this minimalist GP to an extent that it becomes competitive (and in some aspects better) than the state-of-the-art. The use of an e-graph as a support structure for GP brings new light to symbolic regression and GP with the possibility of exploring the accumulated history of the search process and even combining the history of multiple searches.
This paper is organized such that in Section~\ref{sec:relatedwork} we will summarize the related works in symbolic regression. Section~\ref{sec:eqsat} will explain the basic concepts of equality saturation and the e-graph data structure. In Section~\ref{sec:eggGP} we will detail the proposed modifications to the subtree operators. Section~\ref{sec:experiments} and \ref{sec:results} will respectively detail the experiment methods, report and discuss the results. Finally, Section~\ref{sec:conclusions} will provide some final remarks and expectations for the future.

\section{Related work}~\label{sec:relatedwork}
The redundancy of GP search space has been investigated by many authors with conflicting conclusions to whether this is beneficial or not for the search. For example, Ebner~\cite{10.1002/cplx.10021} argued that this redundancy enables the search to reach the optima through different trajectories, increasing the chances of achieving one of the equivalent expressions.
On the other hand, Gustafson et al.~\cite{gustafson2005improving} observed that when the recombination between two similar solutions was forbidden, there was an increase in offsprings that changed the original behavior of their parents, leading to increased performance.

Several works made a detailed study about the redundancy and neutrality in GP (i.e., when a change in the solution has no effect on its outcome). For example, Hu, Banzhaf, and Ochoa \cite{banzhaf2024combinatorics,hu2018neutrality,hu2023phenotype} investigated linear GP for Boolean SR problems with the help of search trajectory networks showing that some phenotypes are overrepresented in the search space. 
Regarding subtree crossover, McPhee et al.~\cite{mcphee2008semantic} showed that over 75\% of crossovers produced no immediately useful semantic changes.


Kronberger et al.~\cite{kronberger2024inefficiency} studied the inefficiency of a simple GP comparing with the enumerated search space~\cite{bartlett2023exhaustive} and using equality saturation to count the percentage of unique expressions generated during the GP search.
They found that from the total of visited expressions during the search, only around $40\%$ were unique. This not only wastes computational resources but it also shows that, at some point, GP fails to explore different regions of the search space.
Many authors observed improvements in the obtained solutions when applying any form of simplification during the search~\cite{cao2023genetic,randall2022bingo,rivero2022dome,seidyo2024inexact} while also stimulating the diversity of the population~\cite{burlacu2019online,burlacu2020hash}.

Equality saturation has been used in the context of symbolic regression as a support tool to study the behavior of the search. Many state-of-the-art SR algorithms have a bias towards creating expressions with redundant numerical parameters~\cite{de2023reducing, kronberger2024jsc}. This redundancy can increase the chance of failing to correctly optimize such parameters, leading to sub-optimal solutions. In~\cite{kronberger2024inefficiency} this technique was used to detect the equivalent expressions visited during the GP search.
So far, the equality saturation technique was not used during the GP search to improve the quality of the solutions.

Semantic similarity is frequently studied in the GP literature, either to improve the population diversity, the locality of the perturbation operators, or to understand the dynamics of GP search~\cite{vanneschi2014survey}. The semantic aware operators~\cite{uy2010semantic} introduce locality by replacing subtrees of an expression with semantically similar trees. Another approach is to combine two expressions $e_1, e_2$ by generating a random expression $e_3$ with a codomain in the range $[0, 1]$ and creating the combined expression $e_ 3 e_1 + (1 - e_3) e_2$, ensuring a balance between the semantics of $e_1$ and $e_2$~\cite{moraglio2012geometric}. In~\cite{ruberto2019genetic}, the authors introduce the \emph{equivalence function} that determines whether two expressions are equivalent if the differences in their behavior in the semantic space is constant. Using this idea, they implement a filtering mechanism that rejects any offspring that is equivalent to any expression in the current population. 

Given the definition of mathematical equivalence, stating that $f = g \iff f(x) = g(x), \forall x \in \mathcal{X}$, where $\mathcal{X}$ is the variable domain, it is important to highlight that in the Semantic GP literature, semantic equivalence is often calculated using a limited number of data points ($\mathcal{X}^* \subset \mathcal{X}$), which cannot guarantee equivalence, but is sufficient for an approximate measure of semantic \textbf{similarity}.
Using equality saturation we can \emph{produce} and \emph{store} equivalent expressions without the need of evaluation, as explained in the next section. 
In this paper, we are concerned with semantic \textbf{equality} and the means to enforce the creation of unvisited expressions, regardless of locality.

\begin{figure}[t!]
    \centering 
    \subfloat[]{\includegraphics[width=0.7\linewidth]{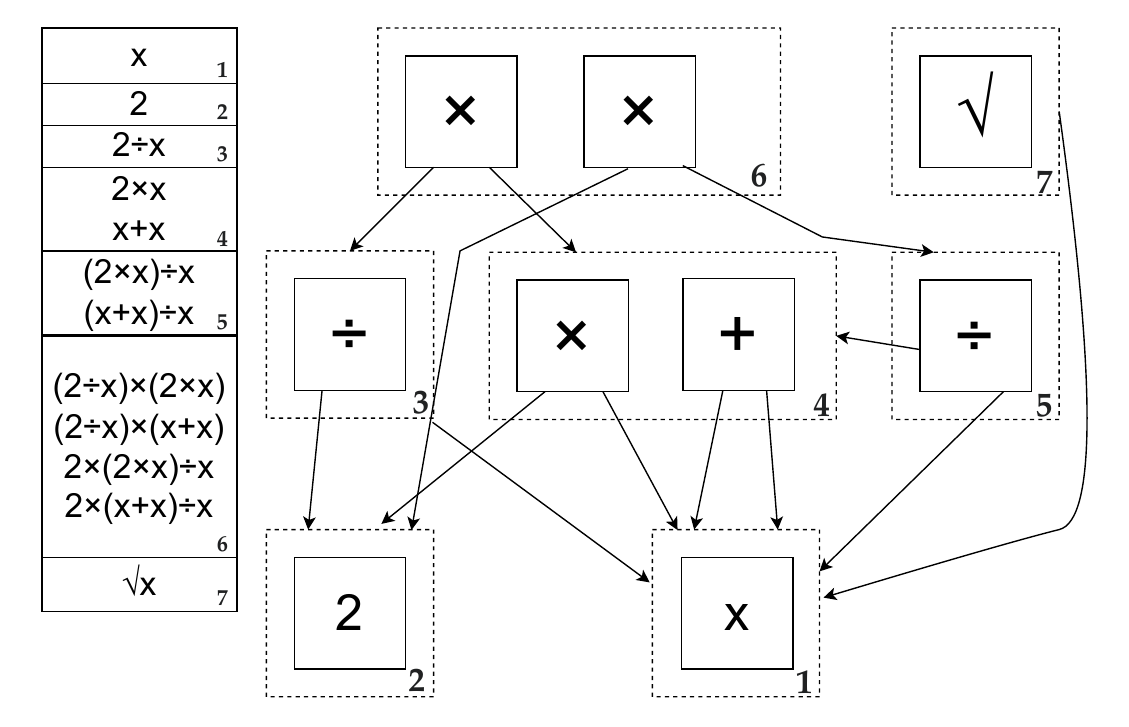}\label{fig:egraph2a}} \\
    \subfloat[]{\includegraphics[trim={1cm 16cm 2cm 1cm},clip,width=0.6\linewidth]{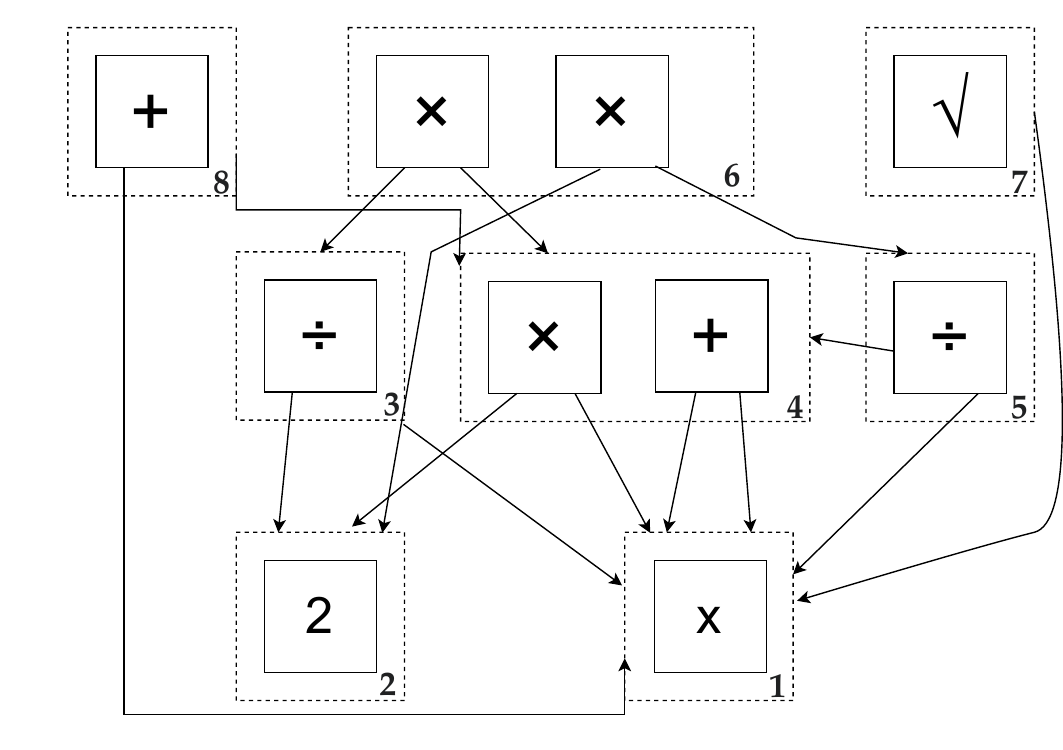}\label{fig:egraph2b}}
    \caption{(a) Illustrative example of an e-graph (the left box shows the expressions evaluated at each e-class) and (b) the same e-graph after inserting the expression $x + 2x$.}
    \label{fig:egraph2}
\end{figure}

\section{Equality saturation and e-graphs}~\label{sec:eqsat}
Equality saturation~\cite{tate2009equality} was proposed as a solution to the phase ordering problem in compiler optimization. This problem occurs when optimizing a program by applying a set of rewrite rules sequentially while dropping the information about the previous versions of the program. If the optimization follows a non-optimal sequence, it will lead to a sub-optimal program. 
Equality saturation solves this issue by applying all of the optimization rules in parallel while keeping the intermediate transformations in a compact form using the data structure called \emph{e-graph}.

Fig.~\ref{fig:egraph2a} illustrates an example of an e-graph. Each solid box represents an e-node that contains a symbol of the expression. The dashed boxes, called e-class, group a set of e-nodes together. Each one of these e-classes is assigned an id (number in the bottom right of an e-class box). The main property of an e-class is that, no matter which e-node is chosen during the traversal, it will lead to an equivalent expression to all other e-nodes of the same e-class. 

Looking at the middle box (e-class id $4$), if we follow through $\times$ it will generate the expression $2x$ and if we follow through $+$ it will generate $x+x$.
The abstract description of the algorithm is very simple, though a concrete and optimal implementation requires the use of advanced techniques and data structures. The main idea is: i) match all the equivalence rules in the current state of the e-graph, ii) apply the rules creating new e-classes, iii) merge the equivalent e-classes, iv) repeat until saturation (i.e., no changes occur).


The data structure of an e-class stores the information about the e-nodes it contains, a list of the parent e-nodes, and additional information also referred to as \emph{semantic analysis}.
This implementation also maintains a database of patterns that allows the algorithm to efficiently match patterns inside the e-graph structure.

The e-graph is commonly used to represent a single program or expression and their equivalent forms, when applying to simplification. But, the structure can keep any number of expressions as long as we keep a list of the e-classes ids that represents the root of each expression. For example, if we insert the expressions $(2/x)(x+x), 2x, \sqrt{x}$ into the e-graph, we would end up with Fig.~\ref{fig:egraph2a}, minus the equivalent relations. As a result, we would keep the list $[\underline{6},\underline{4},\underline{7}]$\footnote{Underlined numbers represent e-class ids.} representing the ids of the expressions we inserted. New expressions are added bottom-up. Starting from a terminal, the algorithm checks whether it already exists in the e-graph returning its e-class id if it does, otherwise, it creates a new id. When adding an internal node, the algorithm first converts it to an e-node by replacing its children by their e-class ids and then it checks whether it already exists in the e-graph, returning the corresponding e-class id or a new one. In our example from Fig.~\ref{fig:egraph2}, if we try to add the expression $x + 2x$, it would first retrieve the e-class ids $\underline{1},\underline{2}$ for the terminals  $x$ and $2$, then it would return the e-class id $4$ corresponding to the e-node $\underline{2} \times \underline{1}$. Finally, it would create a new e-class with id $\underline{8}$ and the e-node $\underline{1} + \underline{4}$. This mechanism allows us to compactly store a set of expressions and readily assert whether an expression already exists in the structure.


\section{\egp: e-graph GP}~\label{sec:eggGP}
The proposed algorithm, \emph{\egp{}} (\textbf{e}-\textbf{g}raph \textbf{g}enetic \textbf{p}rogramming), follows the same structure as the traditional GP. The initial population of size $p$ is created using ramped half-and-half respecting a maximum size and maximum depth~\cite{Koza1992} and, for a number of generations, it will choose $p$ pair of parents using tournament selection, applying the subtree crossover with probability $pc$ followed by the subtree mutation with probability $pm$,  replacing the offsprings following a certain criteria.
The key differences of \egp{} are:
\begin{enumerate}
    \item a single step of equality saturation is executed after inserting new expressions, merging equivalent expressions.
    \item the subtree crossover and mutation are modified to try to generate an unvisited expression.
\end{enumerate}

Notice that a single step of equality saturation will not guarantee the insertion of all equivalent expressions in the e-graph but, if we apply more iterations, the e-graph can grow exponentially large. This issue is amplified by the fact that we are storing multiple expressions. As we will see in Sec.~\ref{sec:results}, the single step seems to be sufficient to improve the results and the benefits of increasing the number of steps is a subject for future research.

We implemented single and multi-objective versions (called \emph{eggp\textsubscript{so}} and \emph{eggp\textsubscript{mo}}), with eggp\textsubscript{so} replacing the current population with the generated offspring and eggp\textsubscript{mo} replacing it by the set of individuals formed by: the Pareto front, the next front after excluding the first Pareto-front, and a selection of the last offspring at random until it reaches the desired population size. Keeping two \emph{ranks} of dominance and filling up the remainder of the population with new expressions is meant to stimulate the combination of new expressions (exploration) while keeping the best fronts (exploitation).
At the end of the execution, we can extract the Pareto-front from the entire history of the search, this is equivalent to the traditional NSGA-II algorithm~\cite{deb2000nsga2} as the dominance relation is transitive.

Moreover, we keep a database of generated expressions sorted by fitness and size (both objectives used in this work), so the Pareto-front can be retrieved in $O(n)$ where $n$ is the number of extracted individuals. A new expression can be inserted into this structure in $O(log(m))$, where $m$ is the number of evaluated expressions so far. Finally, if we add the expression $x+x+x$, equality saturation will generate the equivalent form $\theta x$ (constants are replaced by parameters), storing it in the database of expressions with size $3$.

\subsection{E-graph crossover and mutation}
\begin{figure*}[t!]
\centering
\subfloat[]{\begin{tikzpicture}[scale=0.65,sibling distance=5em,
  every node/.style = {shape=circle,  draw, align=center, top color=white, bottom color=lightgray!20, minimum size=6mm,inner sep=0pt},
  level 1/.style={sibling distance=5em},
  level 2/.style={sibling distance=3em},
    ]]

  \node(first) {$+$}
          child { node  {$\mathbf{x}$} }
          child { node [thick, bottom color=lightgray!80] {$cos$}  
                    child { node {$x$}  }
                }

         ;

  \node(plus)[right= 0.7cm of first, draw=none, fill=none, bottom color=white] {}
       child  { [white] node[draw=none, fill=none, bottom color=white, black] {$+$} };
  
  \node(second)[right= 0.7cm of plus, thick, bottom color=lightgray!80] {$*$}
          child { node  {$+$} 
                    child {  node {$x$} }
                    child {  node [thick, bottom color=lightgray!80]{$2$} }
                }
          child { node {$x$} }

         ;
  \node(equal) [right= 0.7cm of second, draw=none, fill=none, bottom color=white]{}
       child  { [white] node[draw=none, fill=none, bottom color=white, black] {$=$} };

  \node(third)[right= 0.7cm of equal] {$+$}
          child { node  {$\mathbf{x}$} }
          child { node [thick, bottom color=lightgray!80] {$2$}  
                }

         ;
         
\end{tikzpicture}\label{fig:crossover}} \qquad
\subfloat[]{\begin{tikzpicture}[scale=0.65,sibling distance=5em,
  every node/.style = {shape=circle,  draw, align=center, top color=white, bottom color=lightgray!20, minimum size=6mm,inner sep=0pt},
  level 1/.style={sibling distance=5em},
  level 2/.style={sibling distance=3em},
    ]]

  \node(first) {$+$}
          child { node {$x$} }
          child { node [thick, bottom color=lightgray!80] {$\mathbf{cos}$}  
                    child { node {$x$}  }
                }

         ;
 
  \node(equal) [right= 0.7cm of first, draw=none, fill=none, bottom color=white]{}
       child  { [white] node[draw=none, fill=none, bottom color=white, black] {$=>$} }
  ;

  \node(third) [right= 0.7cm of equal] {$+$}
          child { node {$x$} }
          child { node [thick, , bottom color=lightgray!80] {$+$}  
                    child { node [thick, , bottom color=lightgray!80] {$2$} }
                    child { node [thick, , bottom color=lightgray!80] {$x$}  }
                }

         ;
    \node(equal2) [right= 0.7cm of third, draw=none, fill=none, bottom color=white]{}
       child  { [white] node[draw=none, fill=none, bottom color=white, black] {$=>$} }
  ;
    \node(fourth) [right= 0.7cm of equal2] {$+$}
          child { node {$x$} }
          child { node [thick, bottom color=lightgray!80] {$*$}   
                    child { node {$2$} }
                    child { node {$x$}  }
                }

         ;
\end{tikzpicture}\label{fig:mutation}}
\caption{Examples using the e-graph in Fig.~\ref{fig:egraph2b} of (a) recombination between two expressions: after choosing the recombination point marked in bold in the first tree, the second tree has only two points which will generate new expressions (marked in bold in the second expression), after picking one of these points, we generate the new solution illustrated in the tree to the right; (b) mutation: after choosing the mutation point, a new subtree is generated. If the new expression is already contained in the e-graph, the root of the subtree is changed by a random non-terminal that creates an unvisited expression.}

\end{figure*}
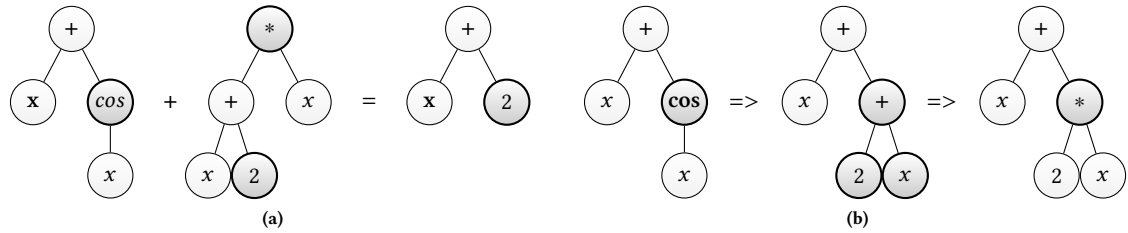

The e-graph crossover and mutation operators exploits the information of the search history stored on the e-graph to increase the probability of generating an unvisited expression.
The e-graph crossover (Fig.~\ref{fig:crossover}) involves two \emph{parents} chosen with tournament selection and replaces a random subtree of the first parent with a random subtree sampled from a subset of all possible subtrees of the second parent. This subset is built such that, when replacing the chosen subtree of the first parent, it will generate an unvisited expression. In the event that this set is empty or the algorithm chooses not to perform the crossover (with probability $1 - pc$), it will return the unmodified first parent.


The e-graph mutation (Fig.~\ref{fig:mutation}) is applied to the offspring of the crossover with a probability $pm$. Like the traditional subtree mutation, it replaces a random subtree of that solution with a randomly generated subtree using either  the grow or full method (chosen at random). After the new expression is created, it is checked whether it already exists in the current e-graph. If it does, the node at the root of the generated subtree is exchanged by another node chosen at random from a subset of the symbols with the same arity. This subset is formed by all the symbols that would create an unvisited expression. When this subset is empty, the current mutated expression is returned.


The full implementation of \egp{} has nine hyperparameters: number of generations, population size, maximum expression size, loss function (MSE, Gaussian, Poisson, Bernoulli, ROXY~\cite{RAR}), number of iterations and retries for the parameter optimization, probabilities of crossover and mutation, and the list of non-terminals.


\section{Experiments}~\label{sec:experiments}
To measure the benefit of stimulating novelty using the history of visited expressions and their equivalent form stored in the e-graph, we chose three baseline algorithms: a version of tinyGP~\cite{Sipper2019tinyGP} implemented using the same backend library, Operon~\cite{burlacu2020operon} and PySR~\cite{cranmerpysr}.

Operon is a carefully crafted implementation of GP for symbolic regression with runtime performance in mind and a good set of default hyperparameters. It incorporates multiple mutation operators that allow a finer perturbation of a solution. Besides, it envelops every variable node with a scaling parameter adjusted using nonlinear optimization. PySR also supports the same mutation operators and nonlinear optimization of the parameters, it stands out with the use of an island model capable of keeping the diversity of the population to stimulate the exploration of the search space. It also applies a simplification heuristic on a selection of the expressions. Both PySR and Operon uses multi-objective optimization with accuracy and expression size as the default objectives.

We should stress that we have kept \egp{} with only the subtree crossover and mutation to be directly comparable with tinyGP, thus measuring the benefits of this modification. 

We have fixed all the common hyperparameters to an empirically set of default values. The only differences in settings are: for PySR we are using $10$ populations in its island model, so the size of each population is $\sfrac{1}{10}$ of the population size for the other algorithms, earlier experiments revealed that PySR performs significantly worse when using a single population; for Operon, we perform a maximum of $100$ optimization iterations instead of $50$ iterations with $2$ different starting points since it does not support multiple restarts; for \egp{}, $\sfrac{1}{3}$ of the training data is separated and used as a validation set to calculate the fitness, while the parameters are fitted using the remaining $\sfrac{2}{3}$. 


\begin{table}[t!]
    \centering
    \caption{Symbolic regression algorithms hyperparameters. Operators enveloped with $|.|$ apply the absolute value to the first argument. The population size for PySR is $\sfrac{1}{10}$ of the reported values in this table to allow the use of ten islands.}
    \label{tab:sr-hyperparameters}
    \begin{tabular}{ll}
    \hline
    \textbf{Parameter}     & \textbf{Value}                                      \\ \hline
    pop. size / gens. / tourn. size  & $500/200/5$                                        \\
    prob. mutation & $0.3$ \\
    prob. crossover & $0.9$ \\ 
    non-terminal set & $+,-,*,\div,\log{(|.|)},\operatorname{exp},\sqrt{|.|},|x|^y)$ \\

    max depth    & $10$                                         \\
    objectives & $[\text{MSE}, \text{size}]$ \\ 
    optimization steps & $2 \times 50$ ($100$ for Operon) \\
    \hline 
    \end{tabular}
\end{table}

For the first set of experiments, we evaluated each algorithm using the recently proposed\footnote{https://github.com/cavalab/srbench/discussions/174\#discussioncomment-10285133} reduced SRBench benchmark. This set is supposed to be representative as it contains datasets with different characteristics. For this experiment, we applied a $3$-fold cross-validation repeating the experiment $10$ times, creating a total of $30$ runs. Finally, we picked some real-world datasets from the literature corresponding to data from different fields, these data already have pre-determined training and test sets, as such we will run $30$ repetitions of each experiment. For every experiment, we store the final Pareto front and will report the performance plot, the area under the curve (AUC), average rank among all datasets, statistical test of the ranks, the average and standard deviation of the running time. 
All algorithms were restricted to a single core to ensure equal conditions.
Table~\ref{tab:datasets} shows the datasets with their corresponding number of data points, features and the chosen maximum size parameter.

\begin{table}[t!]
    \centering
    \caption{Datasets, number of points and variables, and corresponding max. size. Every training set of the SRBench group was capped at $1\,000$ data points chosen at random. For 192\_vineyard we ensured that the rows with $x_0 = x_1 = 0$ were contained in the training set to avoid misbehaving models. The maximum size for Operon is set to  $0.67$ of the maximum size because, internally, Operon will not count the scale  coefficients of a terminal towards the model size. This factor enforces Operon to search on a similar search space as the other algorithms. These values are in parentheses.}
    \label{tab:datasets}
    \begin{tabular}{lrrrr}
    \hline
    \textbf{Name} & \textbf{Points} & \textbf{Features} & \textbf{max. size} \\
    \hline 
    \multicolumn{4}{l}{SRBench} \\
    \hline 
    192\_vineyard & $52$ & $2$ & $50 (33)$ \\
    210\_cloud & $108$ & $5$ & $50 (33)$ \\
    522\_pm10 & $500$ & $7$ & $50 (33)$ \\
    557\_analcatdata\_apnea1 & $475$ & $3$ & $50 (33)$ \\
    579\_fri\_c0\_250\_5 & $250$ & $5$ & $50 (33)$ \\
    606\_fri\_c2\_1000\_10 & $1\,000$ & $10$ & $50 (33)$ \\
    650\_fri\_c0\_500\_50 & $500$ & $50$ & $50 (33)$ \\
    678\_visualizing\_environmental & $111$ & $3$ & $50 (33)$ \\
    1028\_SWD & $1\,000$ & $10$ & $50 (33)$ \\
    1089\_USCrime & $47$ & $13$ & $50 (33)$ \\
    1193\_BNG\_lowbwt & $31\,104$ & $9$ & $50 (33)$ \\
    1199\_BNG\_echoMonths & $17\,496$ & $9$ & $50 (33)$ \\
    \hline
    \multicolumn{4}{l}{Real world} \\
    \hline
    Chemical\_1\_tower & $4\,999$ & $25$ & $30 (20)$  \\ 
    Chemical\_2\_competition & $1\,066$ & $57$ & $30 (20)$  \\ 
    Friction\_stat\_one-hot & $2\,016$ & $16$ & $30 (20)$ \\ 
    Friction\_dyn\_one-hot & $2\,016$ & $17$ & $30 (20)$  \\ 
    Flow\_stress\_phip0.1 & $7\,800$ & $2$ & $20 (13)$  \\
    Nasa\_battery\_1\_10min & $636$  &$6$ & $20 (13)$  \\
    Nasa\_battery\_2\_20min & $1\,638$ & $5$ & $20 (13)$  \\
    Nikuradse\_1 & $362$ & $2$ & $20 (13)$  \\ 
    Nikuradse\_2 & $362$ & $1$ & $20 (13)$ \\
    \hline 
    \end{tabular}
\end{table}

\section{Results and Discussion}~\label{sec:results}
\begin{figure*}%
\centering
\subfloat[192]{\includegraphics[clip,width=0.22\linewidth]{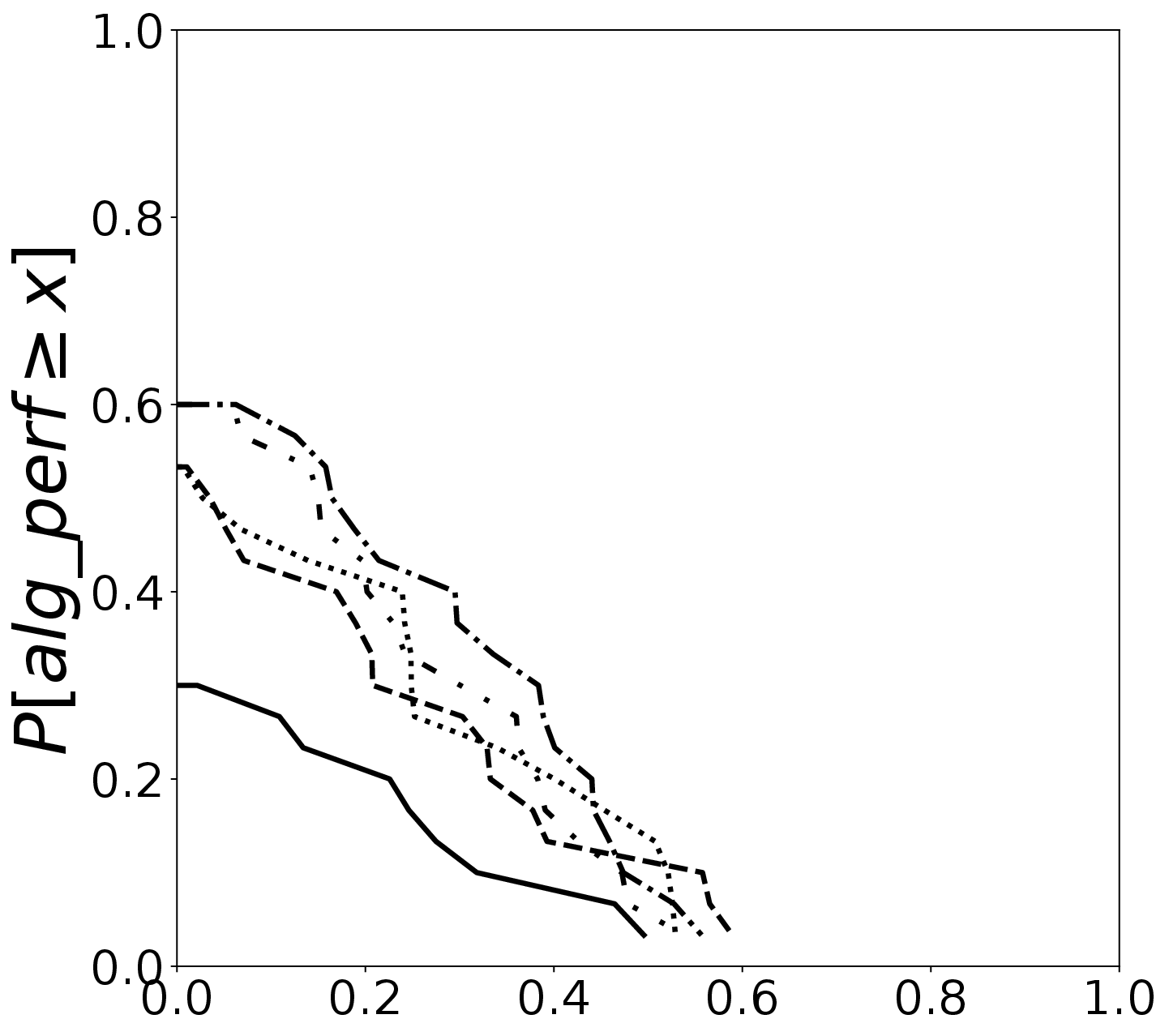}} 
\subfloat[210]{\includegraphics[width=0.2\linewidth]{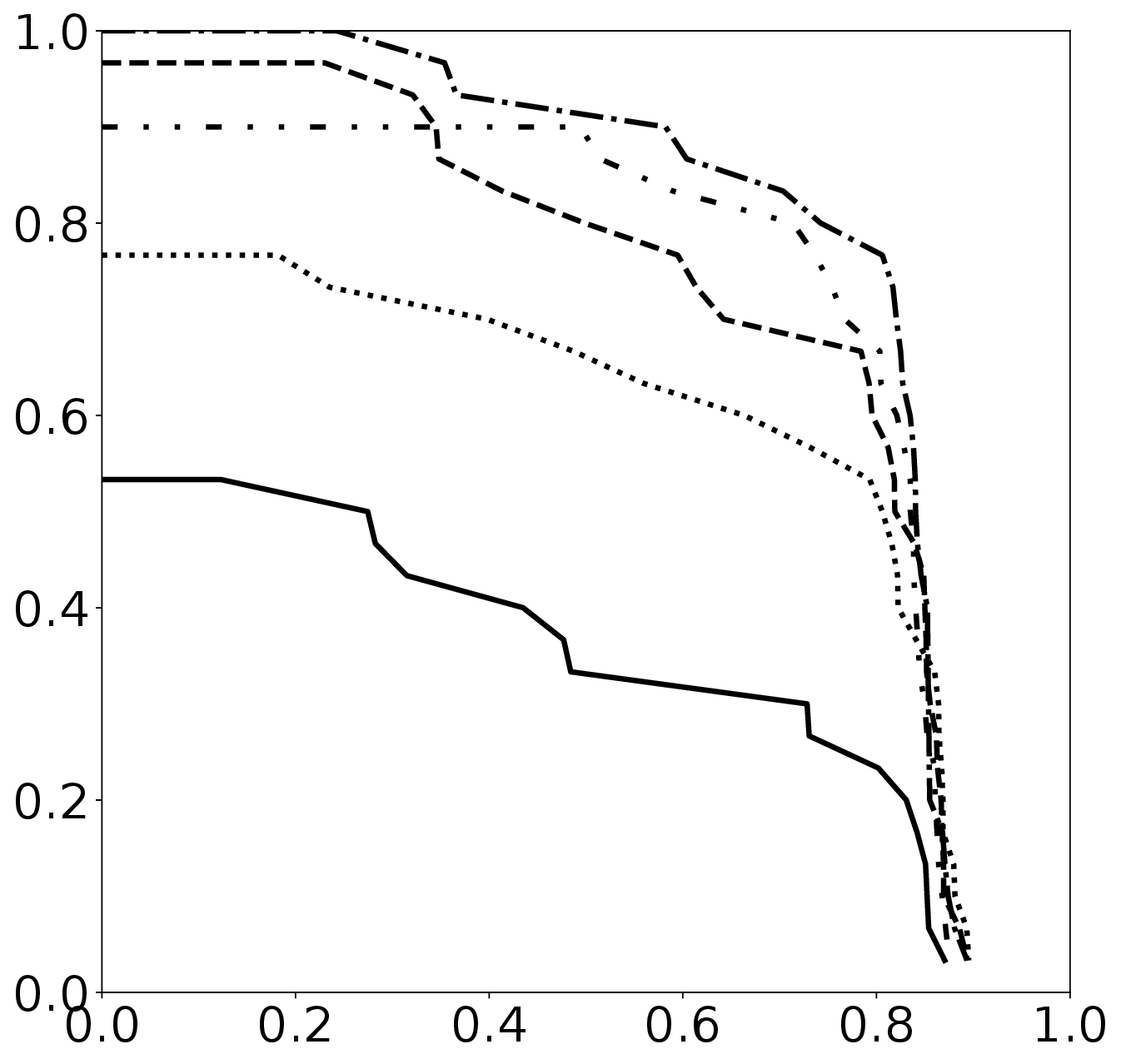}} 
\subfloat[522]{\includegraphics[width=0.2\linewidth]{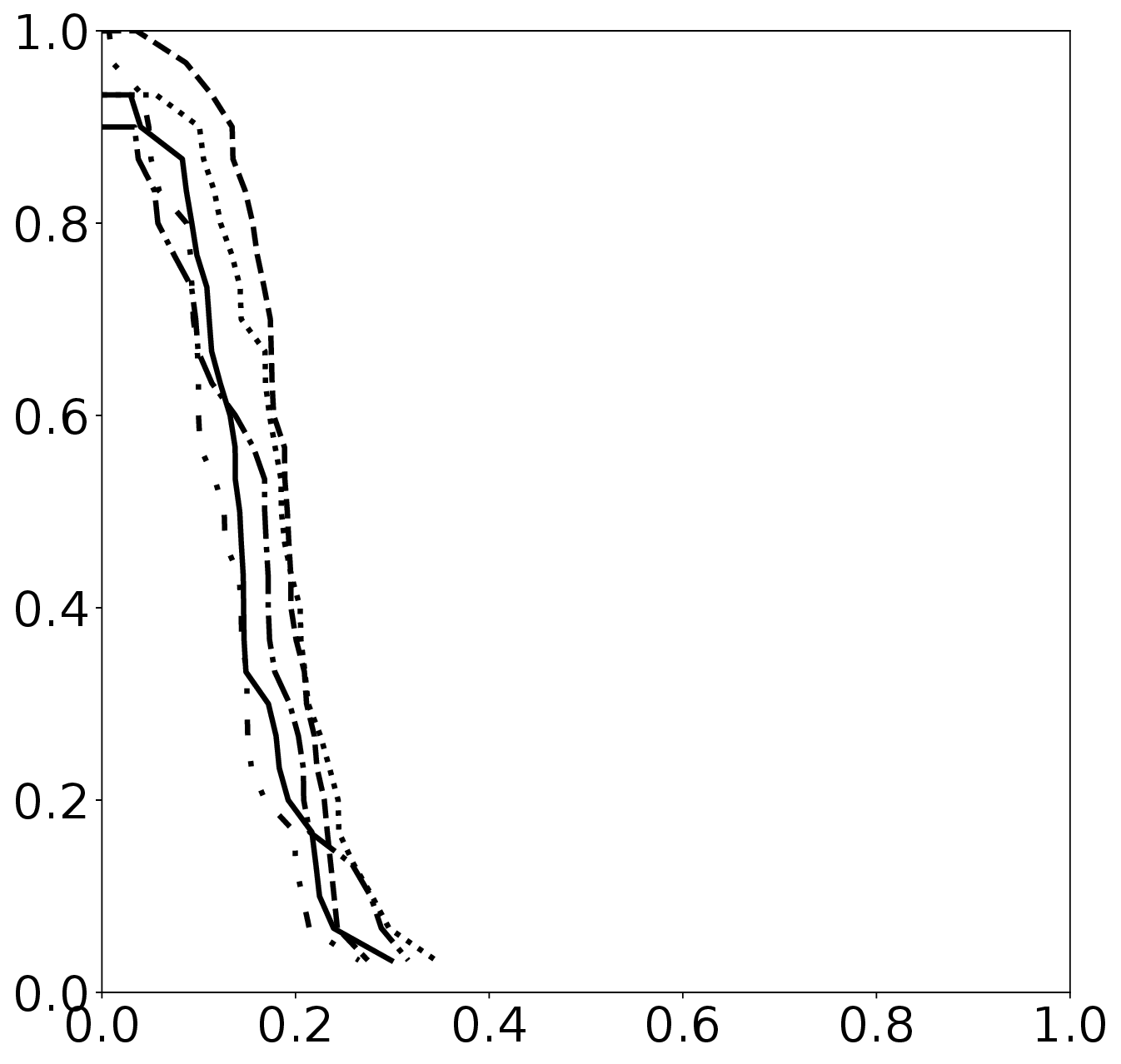}} 
\subfloat[557]{\includegraphics[width=0.2\linewidth]{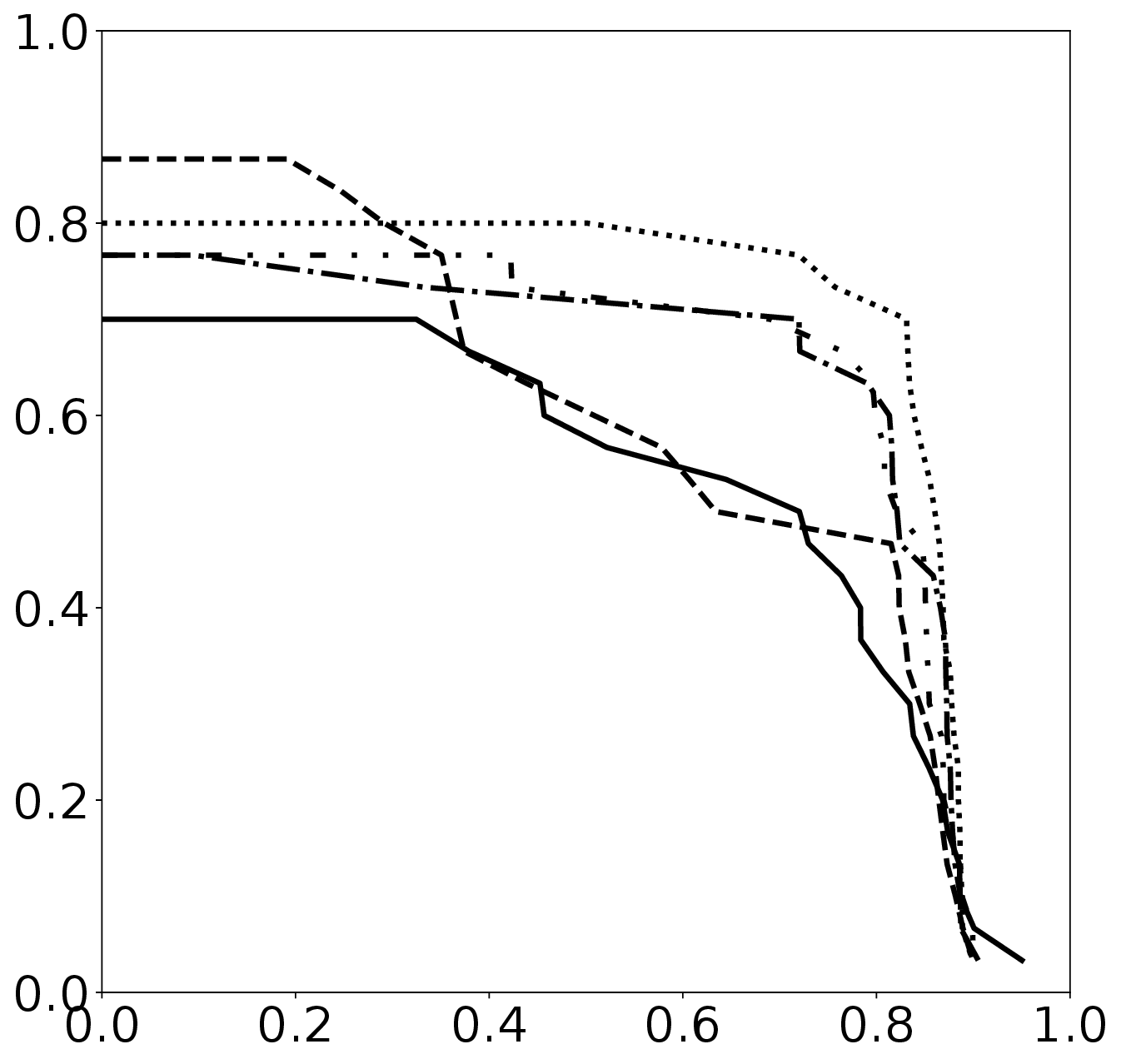}}
\\
\subfloat[579]{\includegraphics[width=0.22\linewidth]{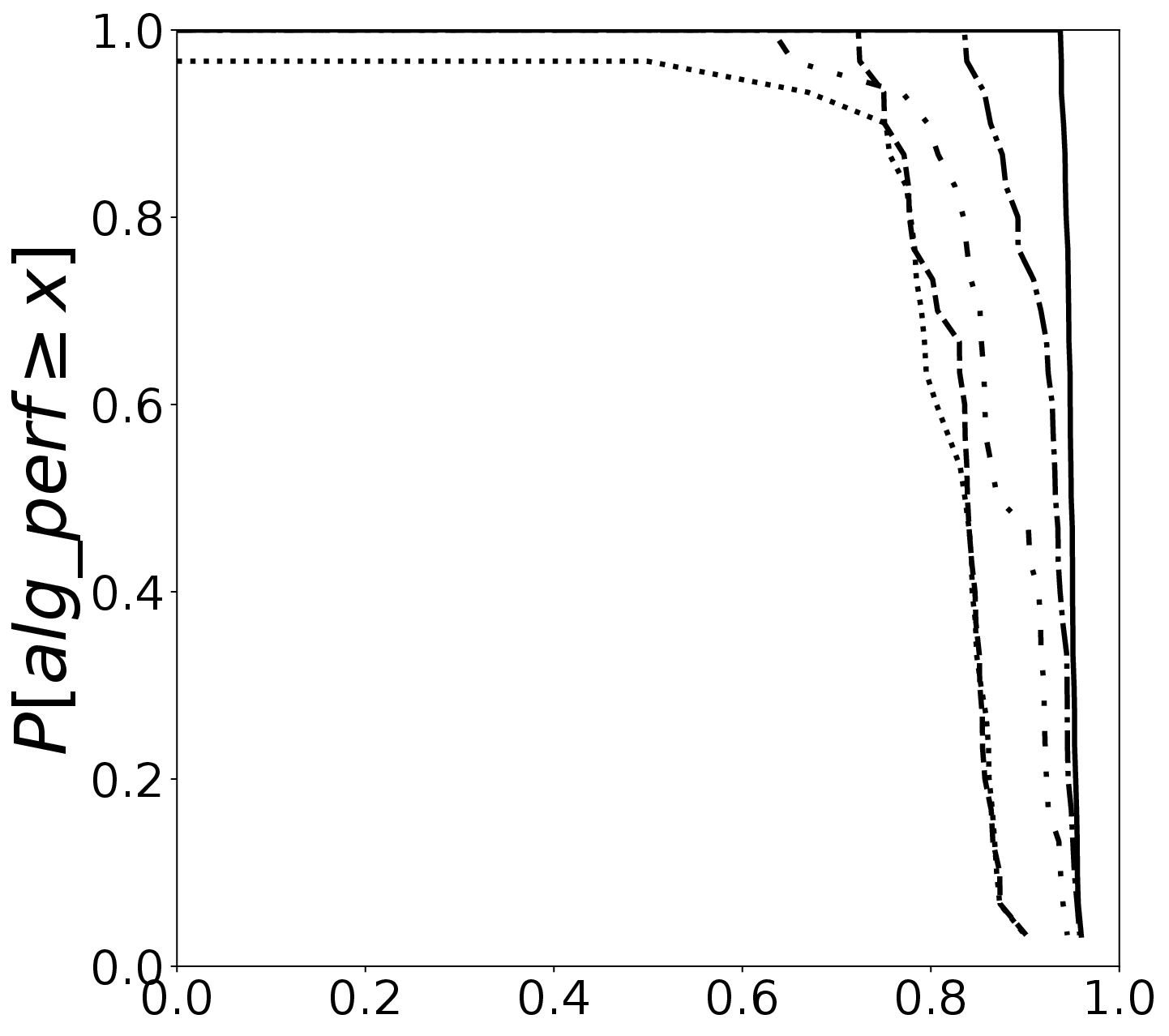}} 
\subfloat[606]{\includegraphics[width=0.2\linewidth]{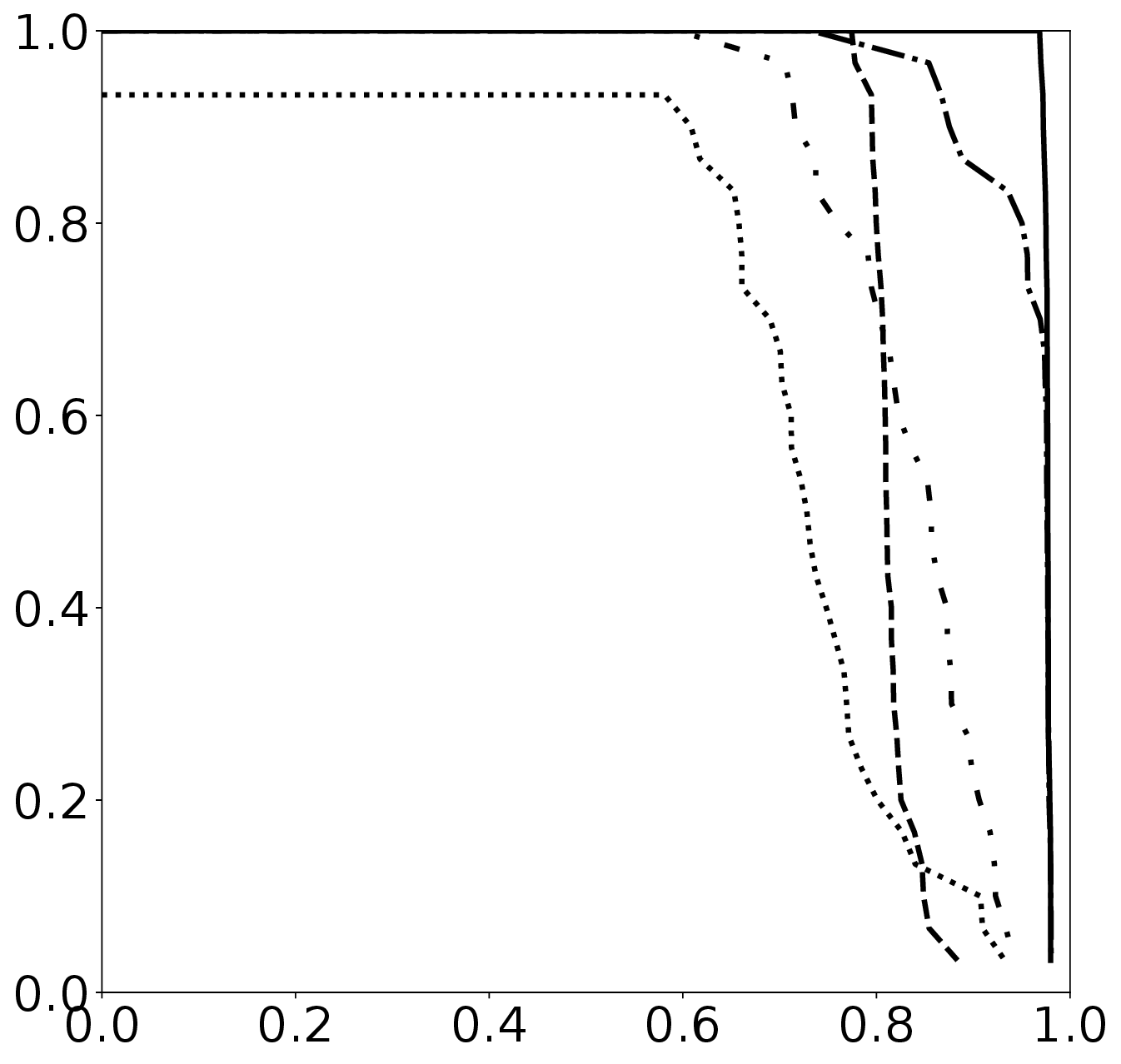}} 
\subfloat[650]{\includegraphics[width=0.2\linewidth]{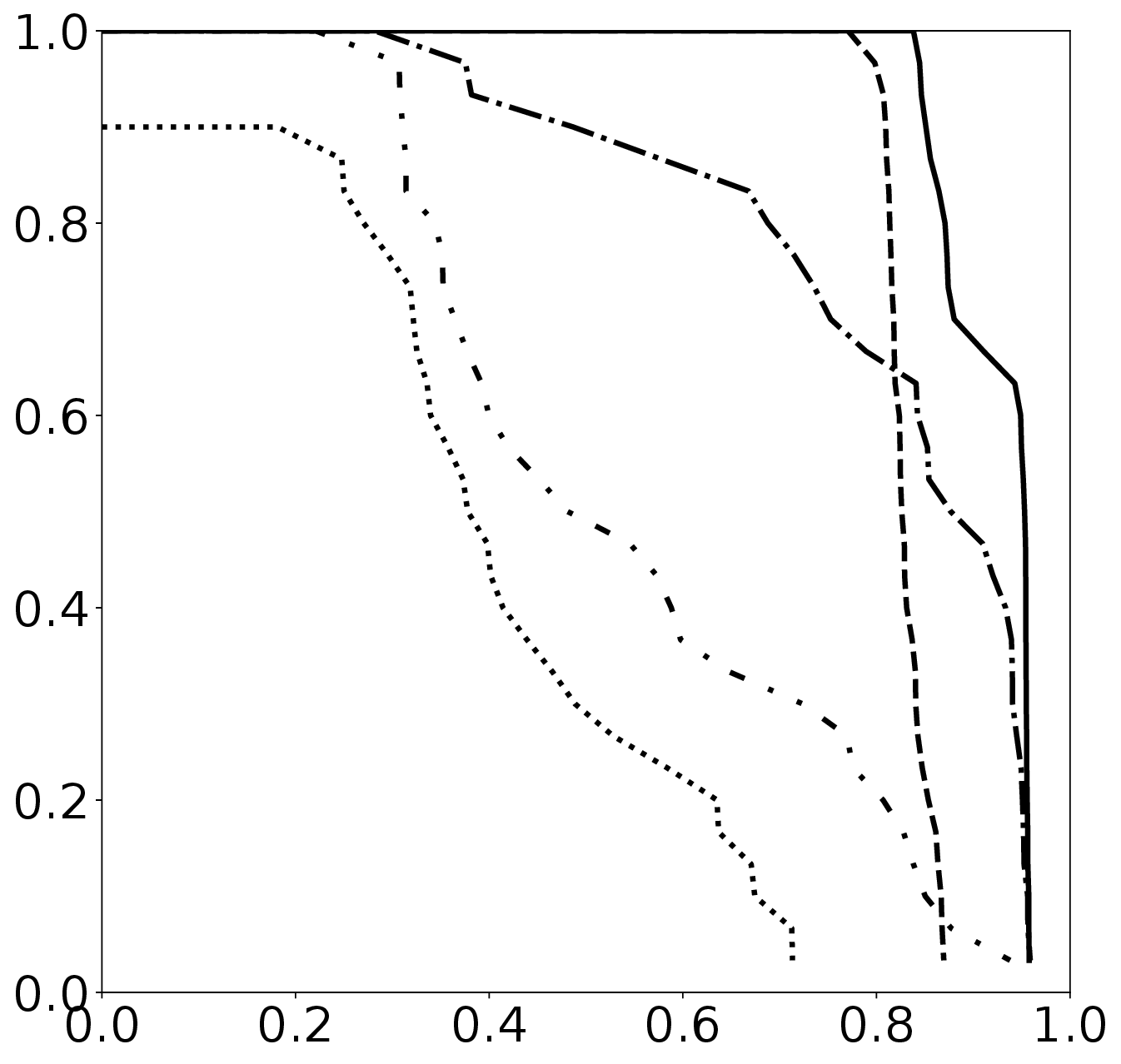}} 
\subfloat[678]{\includegraphics[width=0.2\linewidth]{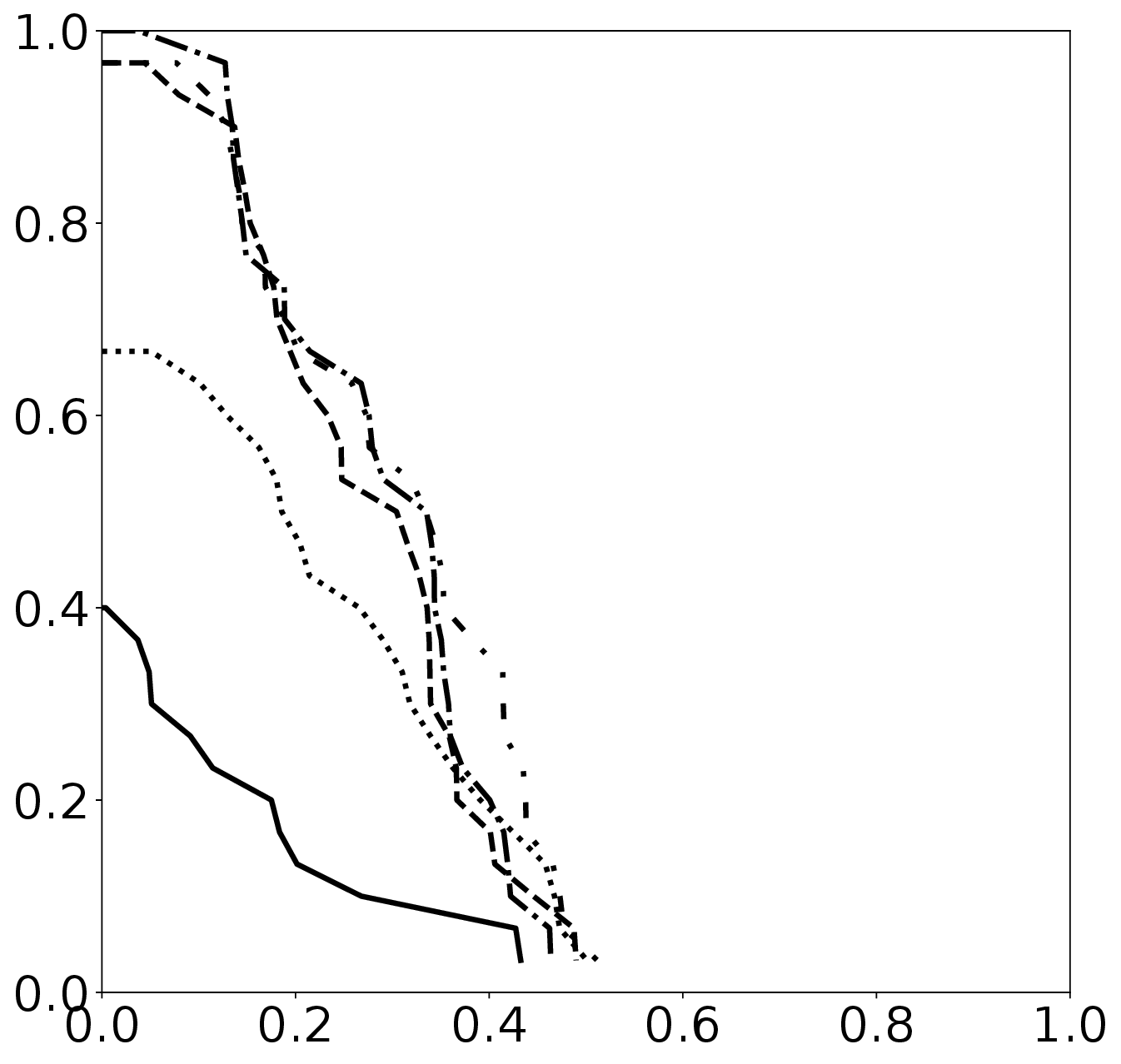}}
\\
\subfloat[1028]{\includegraphics[width=0.22\linewidth]{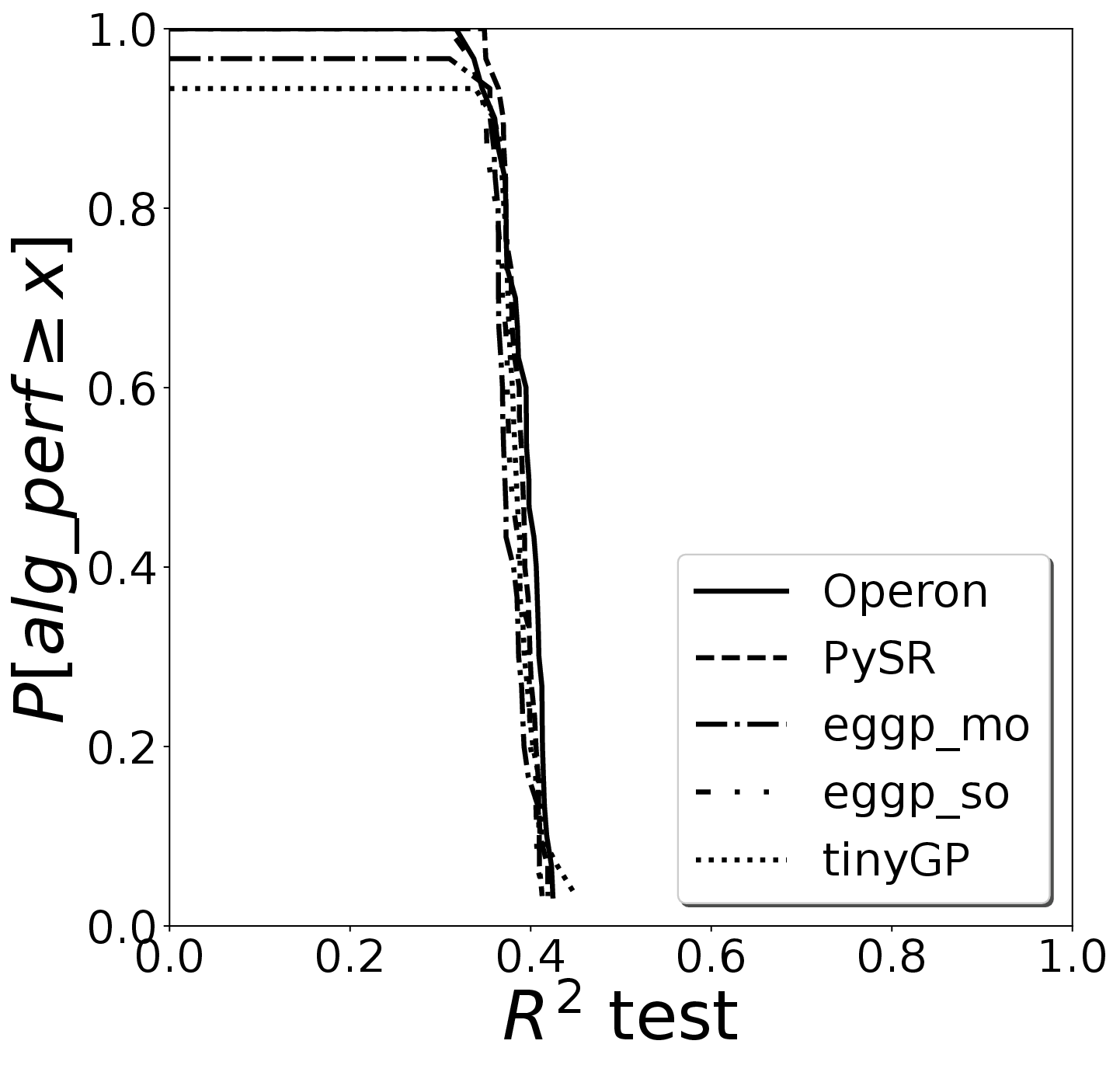}} 
\subfloat[1089]{\includegraphics[width=0.2\linewidth]{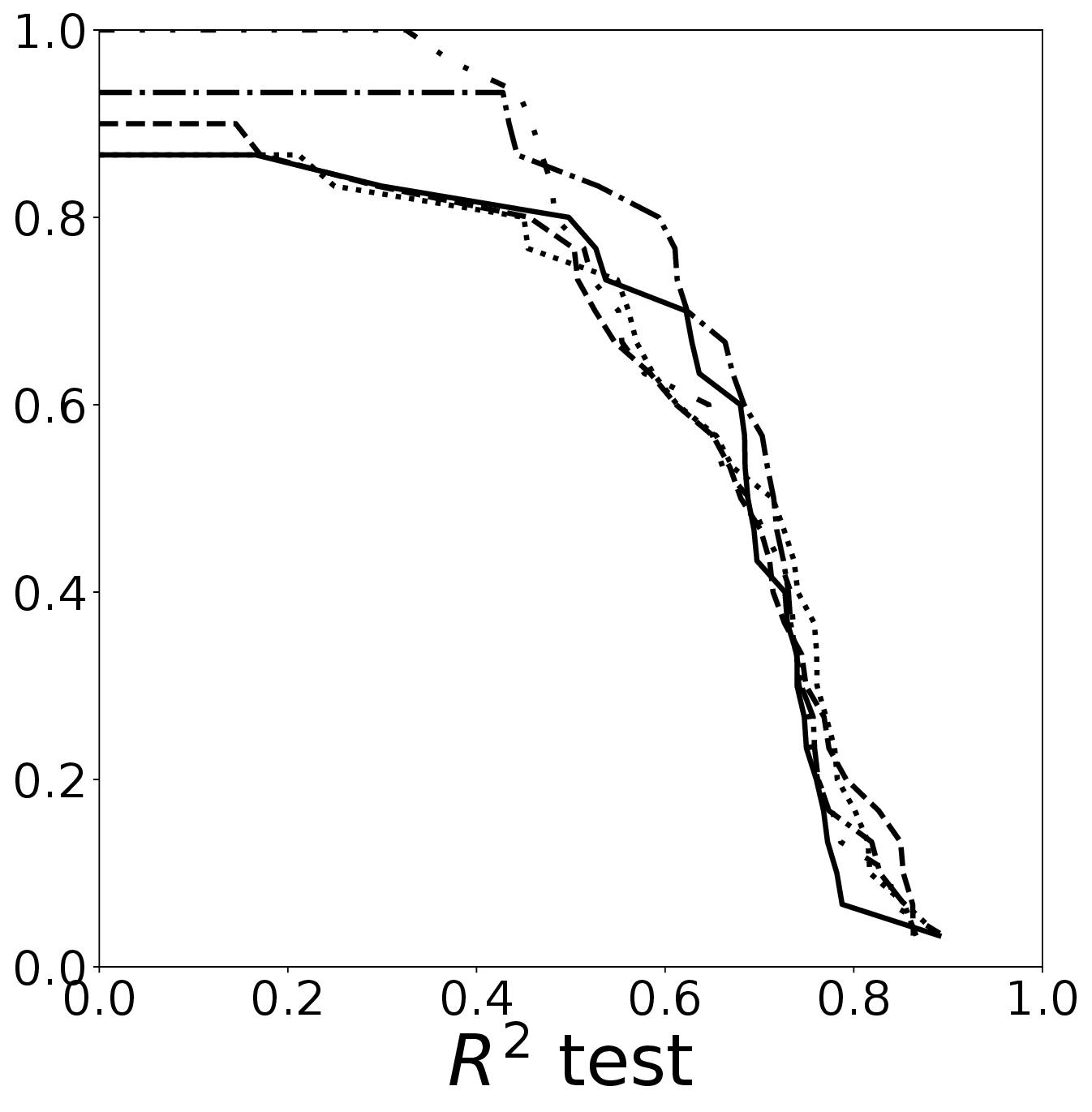}} 
\subfloat[1193]{\includegraphics[width=0.2\linewidth]{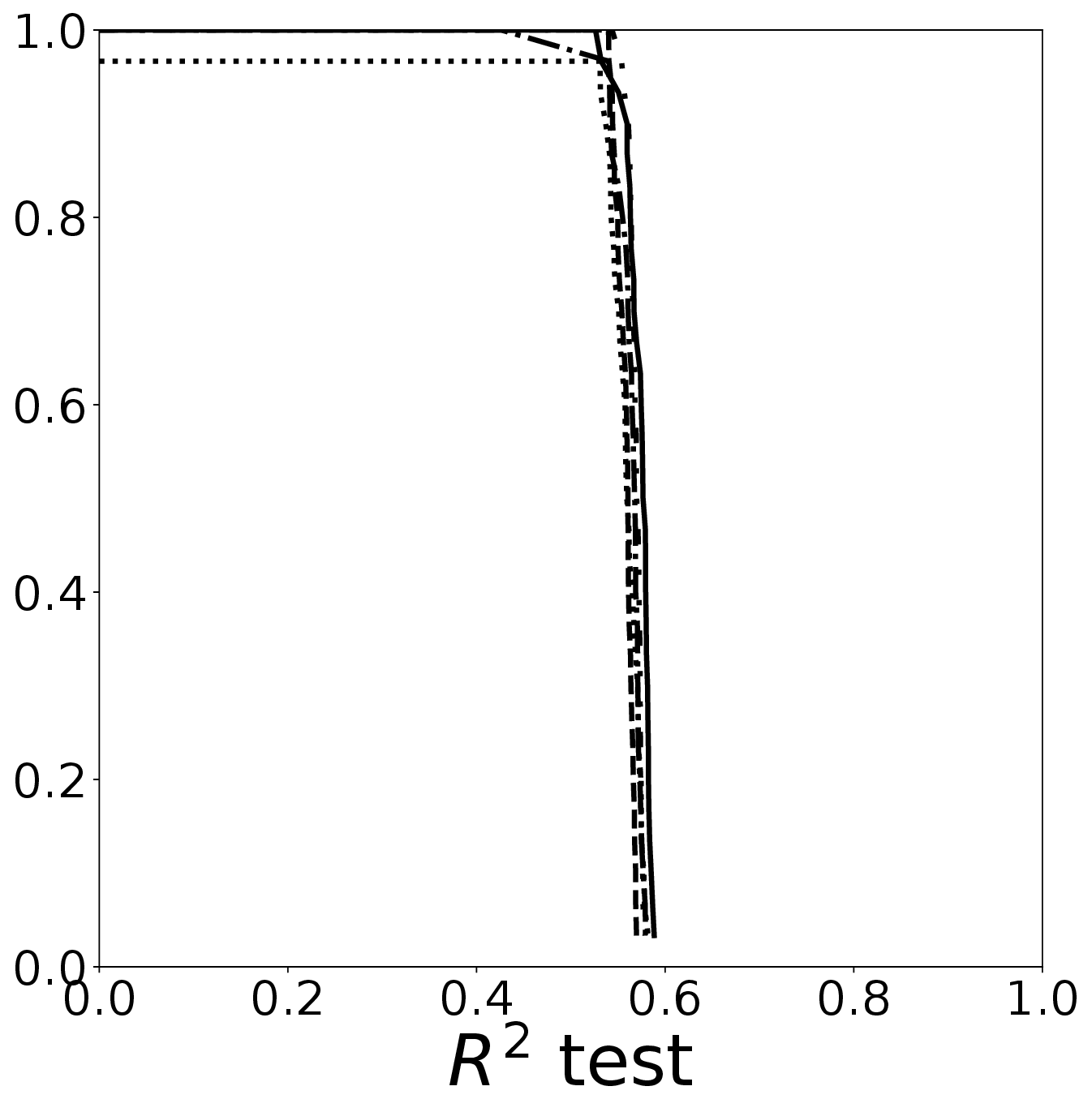}} 
\subfloat[1199]{\includegraphics[width=0.2\linewidth]{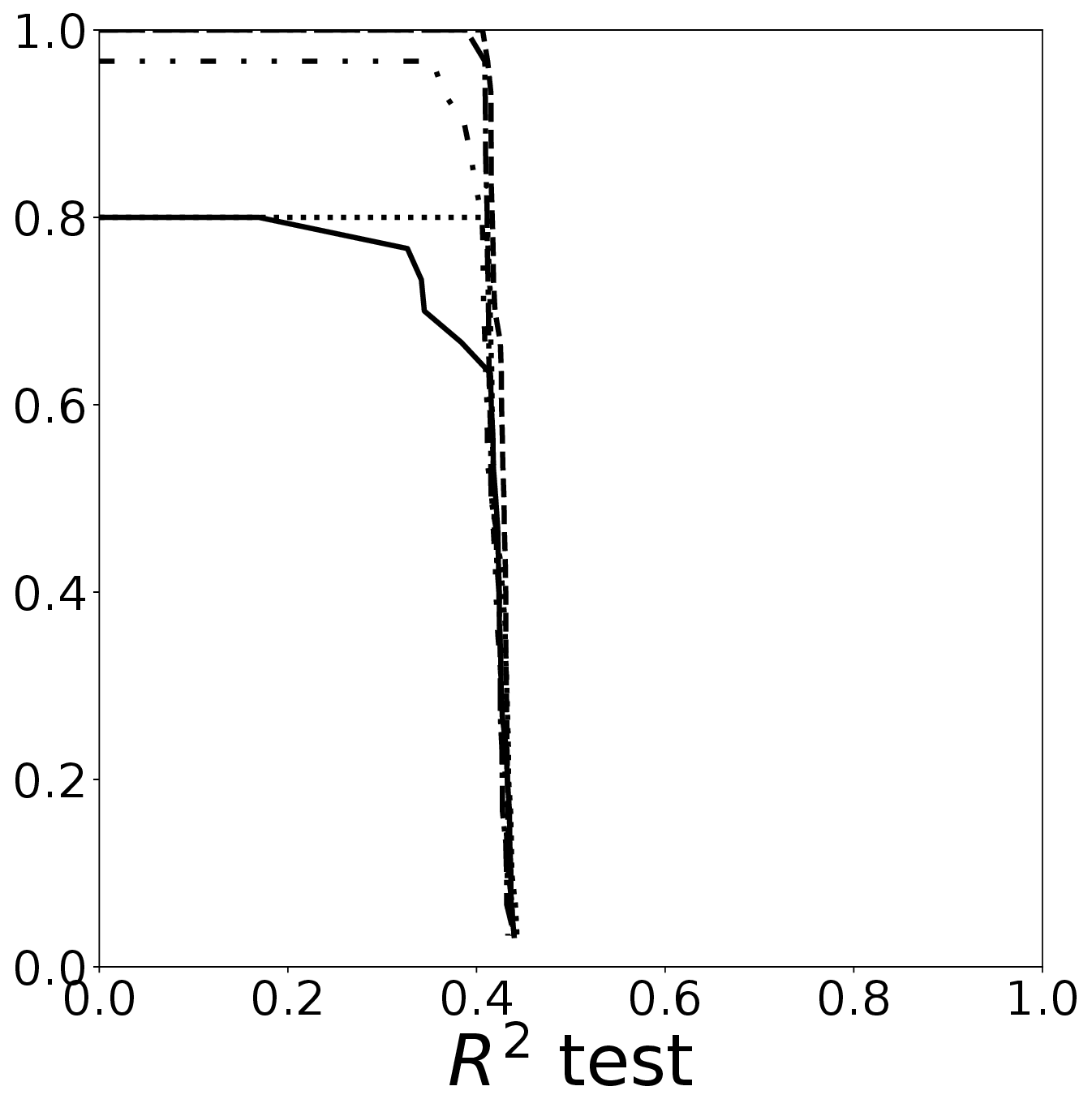}}
\caption{Performance plots for the SRBench datasets. This plot shows the probability of returning an $R^2$ equal or larger than $x$ on a random run of each algorithm.}
\label{fig:srbench-results}
\end{figure*}

In Fig.~\ref{fig:srbench-results} we can see the performance plots for the \textbf{SRBench datasets} considering the best solution of each run according to the highest $R^2$ on the training set. The x-axis of these plots represents the $R^2$ measured on the test set, and the y-axis shows the percentage of runs that the algorithm found an $R^2$ equal or larger than $x$. The ideal algorithm would cover the whole area from $(0,0)$ to $(1,1)$.
From these plots we can see that in at least $3$ datasets ($522$, $1028$ and $1193$) every algorithm reliably achieve the same $R^2$. In other datasets ($579$, $606$, $1089$) both versions of \egp{} maintain this reliability (i.e., achieves the same score in almost every execution) while the other algorithms either achieve a lower score or fails in some execution. The failing executions can be identified as $1 - P(R^2 >0)$, for dataset $557$ every algorithm fails between $15\%$ to $35\%$ of the times. 

In Table~\ref{tab:rank-srbench} we can see the ranks when considering the median $R^2$ of the test set and the AUC. Using this criteria, \egp{}\textsubscript{mo} have an average rank of $2.42$, while Operon comes next ranked $2.5$ on average. Considering \egp{}\textsubscript{so} and tinyGP, there is a slightly decrease in the average rank when using the proposed operators. The statistical test reveals that we can reject the null hypothesis when comparing to PySR with the alternative of having greater median rank. On the other hand, for the AUC values, we can see that \egp{}\textsubscript{mo} is greater than the other algorithms on average while rejecting the null hypotheses for each comparison, except Operon. Unlike the median of the $R^2$, the AUC is the average $R^2$ weighted by the probability of obtaining that value or greater, acting as a reliability measure.
Also in this table, we can see that, on average, \egp{} (both versions) consistently return smaller models than the competing algorithms. There are two possible reasons for this behavior: i) as we apply equality saturation after inserting each expression into the e-graph, they can result in a simplified version of the inserted expression, ii) during the insertion, it automatically eliminates some of the redundant parameters, avoiding the issue reported in\cite{de2023reducing}.

\begin{table}[t!]
\centering
\caption{Ranks of the median (1st block), AUC (2nd block), and average size (3rd block) of the test set $R^2$ for the SRBench. The $p$-values were calculated with a Wilcoxon signed-rank test using as alternative hypotheses ($\alpha = 0.05$) being greater ($>$) than \egp{}.}\label{tab:rank-srbench}
\begin{tabular}{lRRRRR}
\toprule
dataset & \text{eggp}\textsubscript{mo} & \text{eggp}\textsubscript{so} & \text{Operon} & \text{PySR} & \text{tinyGP} \\
\midrule
192 & \mathbf{1} & 2 & 5 & 4 & 3 \\
210 & 2 & \mathbf{1} & 5 & 3 & 4 \\
522 & 3 & 5 & 4 & \mathbf{1} & 2 \\
557 & 2 & 3 & 5 & 4 & \mathbf{1} \\
579 & 2 & 3 & \mathbf{1} & 4 & 5 \\
606 & \mathbf{1} & 3 & 2 & 4 & 5 \\
650 & 2 & 4 & \mathbf{1} & 3 & 5 \\
678 & 2 & \mathbf{1} & 5 & 3 & 4 \\
1028 & 5 & 4 & \mathbf{1} & 2 & 3 \\
1089 & 3 & 4 & 2 & 5 & \mathbf{1} \\
1193 & 3 & 2 & \mathbf{1} & 4 & 5 \\
1199 & 4 & 5 & 2 & \mathbf{1} & 3 \\
\midrule
mean & \mathbf{2.50} & 3.08 & 2.83 & 3.17 & 3.42 \\
$p$-value $>$ & & $0.05$ & $0.21$ & $0.02$ & $0.11$ \\
\midrule \midrule
192 & \mathbf{0.22} & 0.18 & 0.08 & 0.16 & 0.17 \\
210 & \mathbf{0.75} & 0.72 & 0.42 & 0.71 & 0.58 \\
522 & 0.15 & 0.12 & 0.15 & \mathbf{0.19} & 0.18 \\
557 & 0.62 & 0.65 & 0.48 & 0.58 & \mathbf{0.68} \\
579 & 0.92 & 0.87 & \mathbf{0.95} & 0.83 & 0.79 \\
606 & 0.96 & 0.84 & \mathbf{0.97} & 0.82 & 0.70 \\
650 & 0.81 & 0.55 & \mathbf{0.92} & 0.83 & 0.38 \\
678 & 0.29 & \mathbf{0.30} & 0.07 & 0.27 & 0.20 \\
1028 & 0.36 & 0.38 & \mathbf{0.39} & 0.39 & 0.36 \\
1089 & 0.65 & 0.65 & \mathbf{0.66} & 0.59 & 0.59 \\
1193 & 0.56 & 0.57 & \mathbf{0.57} & 0.56 & 0.54 \\
1199 & 0.42 & 0.40 & 0.31 & \mathbf{0.43} & 0.35 \\
\midrule 
mean & \mathbf{0.56} & 0.52 & 0.50 & 0.53 & 0.46 \\
$p$-value $>$ & & $0.05$ & $0.23$ & $0.05$ & $0.01$ \\
\midrule\midrule
192 & \mathbf{17.27} & 28.40 & 48.87 & 44.57 & 48.76 \\
210 & \mathbf{20.73} & 25.20 & 44.67 & 43.50 & 48.70 \\
522 & \mathbf{29.23} & 31.77 & 48.03 & 34.23 & 49.14 \\
557 & 32.03 & 35.07 & 47.07 & \mathbf{25.80} & 49.00 \\
579 & \mathbf{38.97} & 42.63 & 49.13 & 40.93 & 49.00 \\
606 & \mathbf{40.80} & 41.33 & 48.70 & 41.17 & 48.43 \\
650 & 33.90 & \mathbf{22.30} & 48.73 & 42.90 & 47.81 \\
678 & \mathbf{14.57} & 19.37 & 48.53 & 43.63 & 48.68 \\
1028 & 38.03 & 39.60 & 48.80 & \mathbf{37.63} & 49.18 \\
1089 & \mathbf{16.83} & 19.93 & 49.27 & 37.23 & 49.04 \\
1193 & \mathbf{30.57} & 36.33 & 48.63 & 31.90 & 49.07 \\
1199 & \mathbf{24.03} & 29.03 & 47.60 & 37.27 & 48.89 \\
\midrule
mean & \mathbf{28.08} & 30.91 & 48.17 & 38.40 & 48.81 \\
\bottomrule
\end{tabular}
\end{table}

\begin{figure*}%
\centering
\subfloat[chemical 1]{\includegraphics[width=0.2\linewidth]{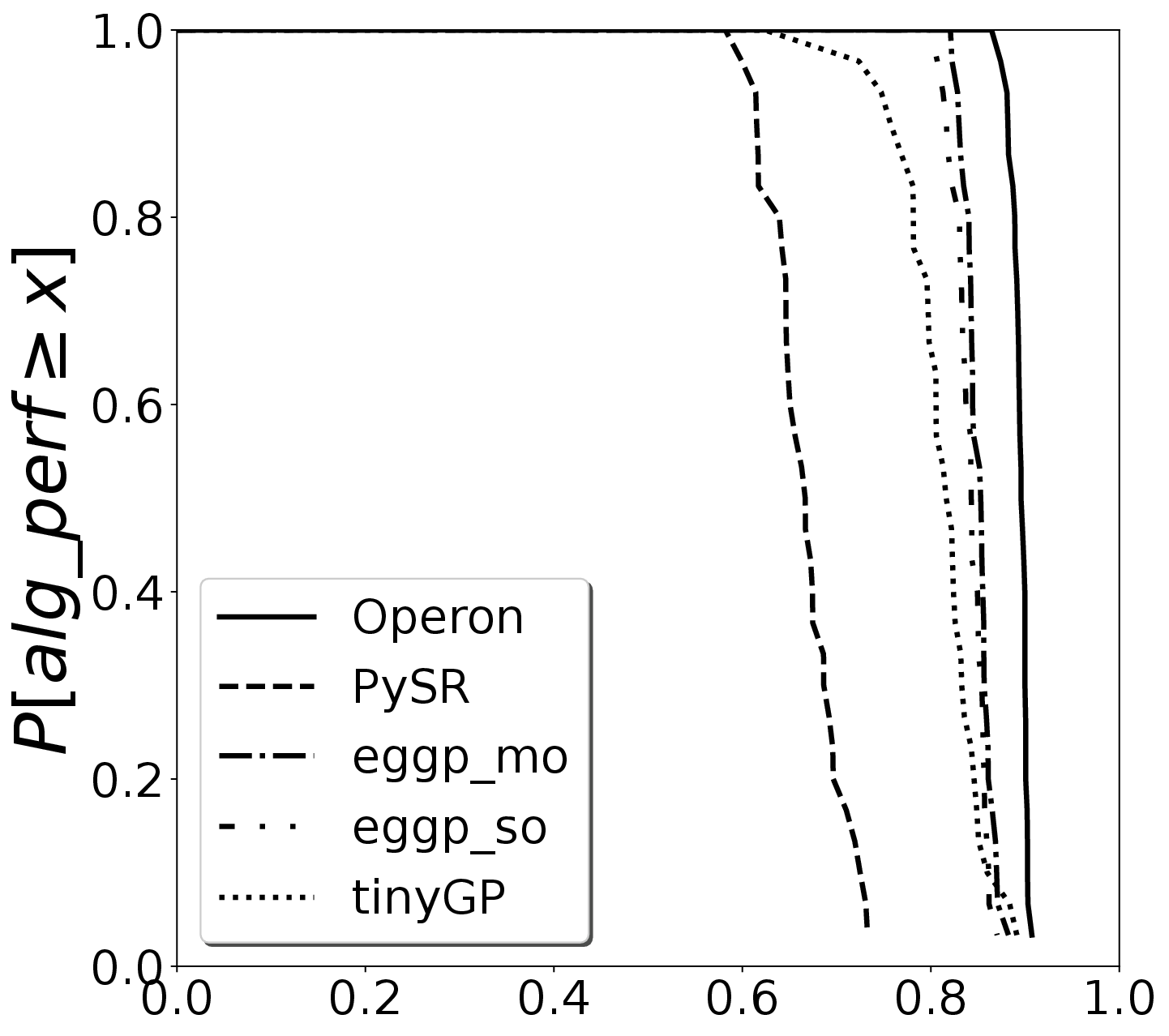}} 
\subfloat[chemical 2]{\includegraphics[width=0.19\linewidth]{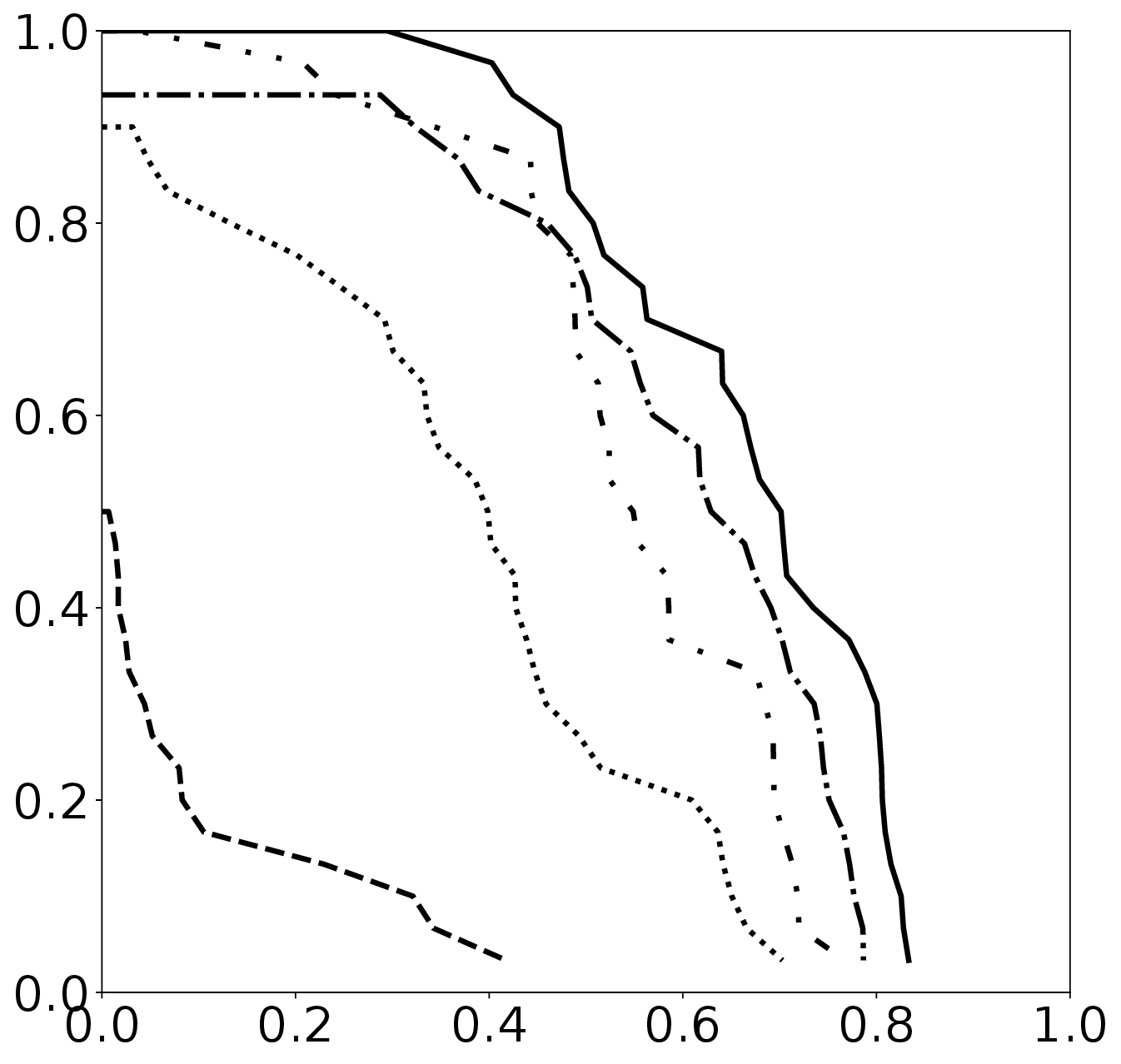}} 
\subfloat[flow]{\includegraphics[width=0.19\linewidth]{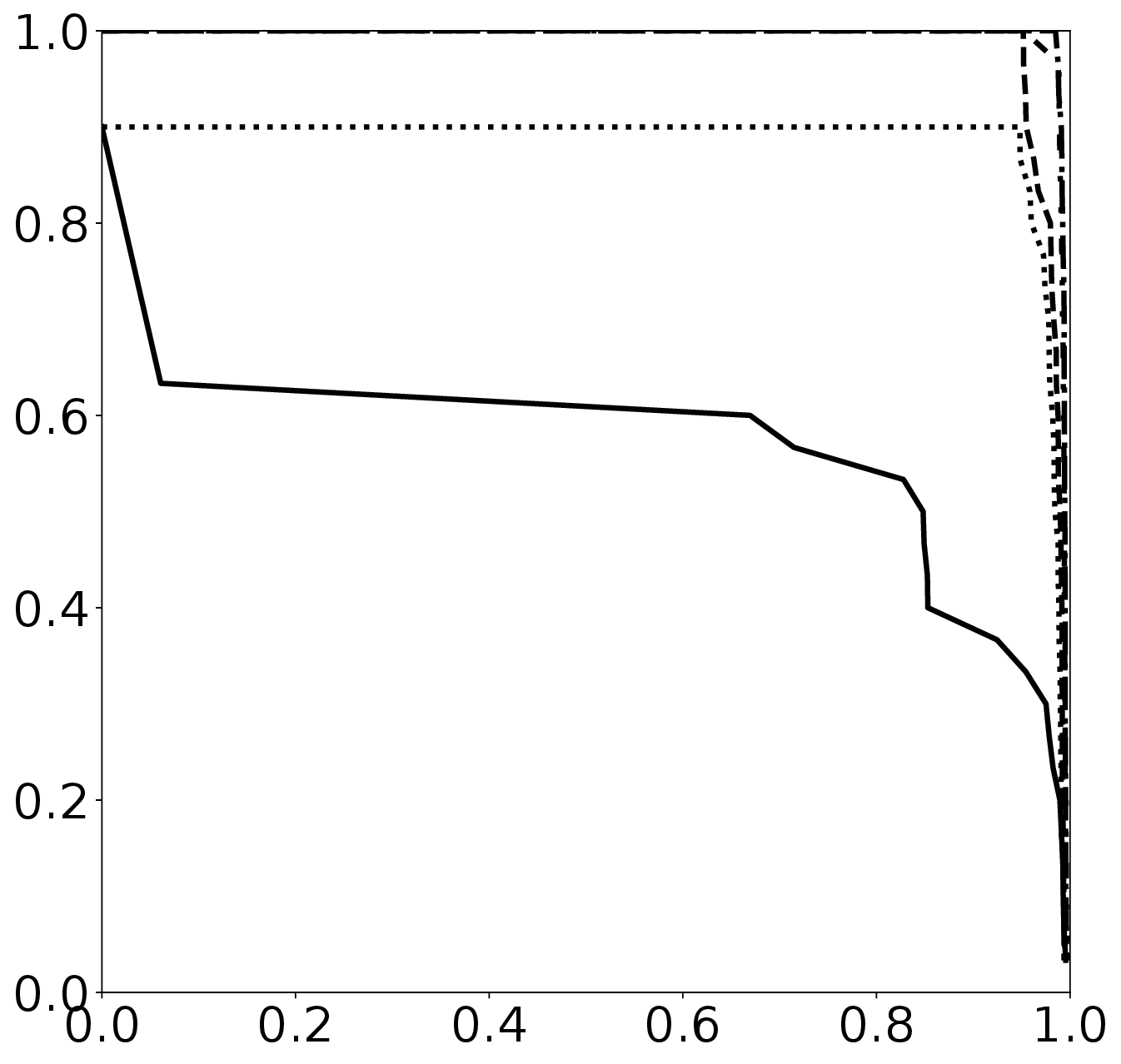}}
\subfloat[friction]{\includegraphics[width=0.19\linewidth]{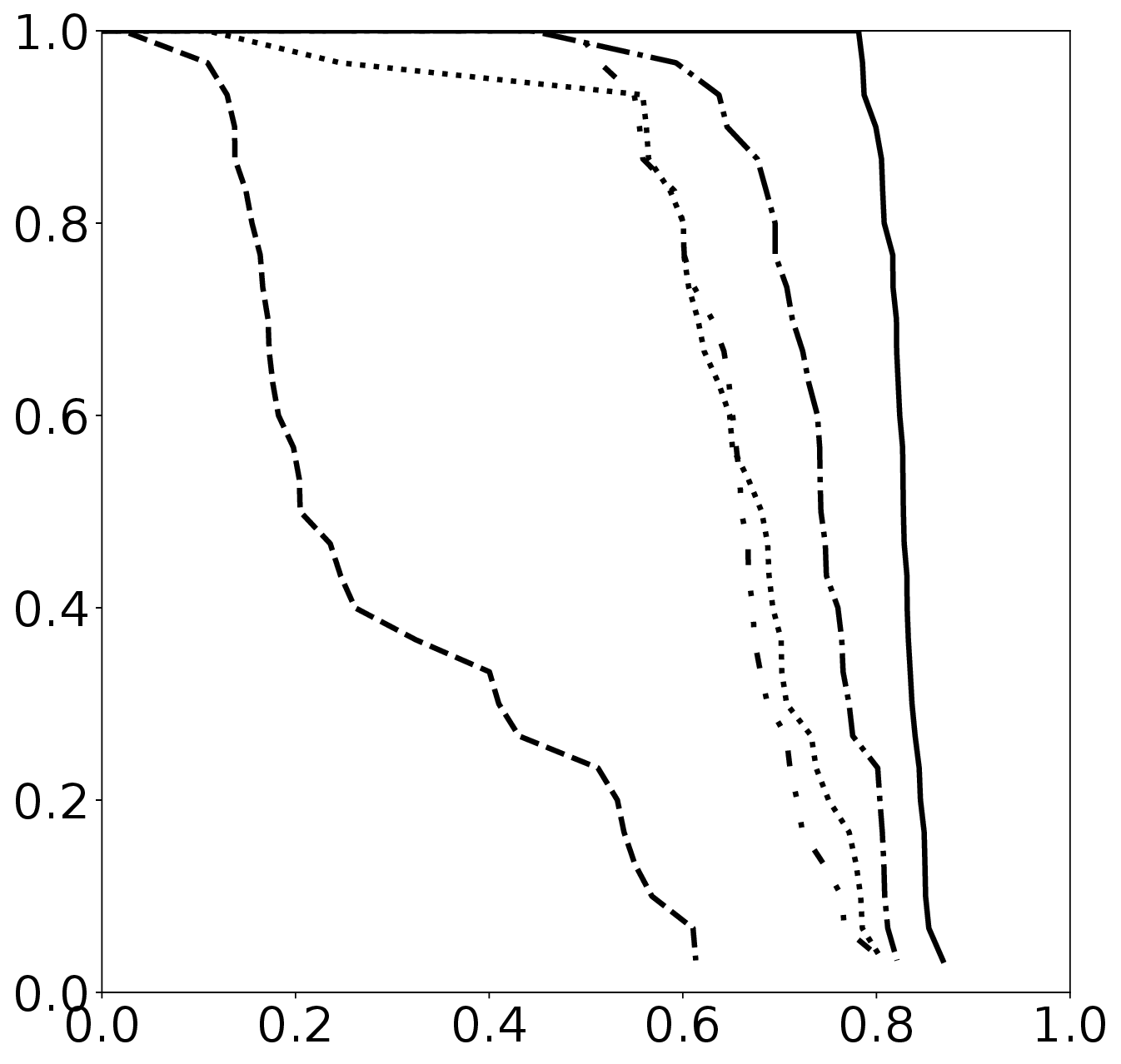}}
\subfloat[friction dyn]{\includegraphics[width=0.19\linewidth]{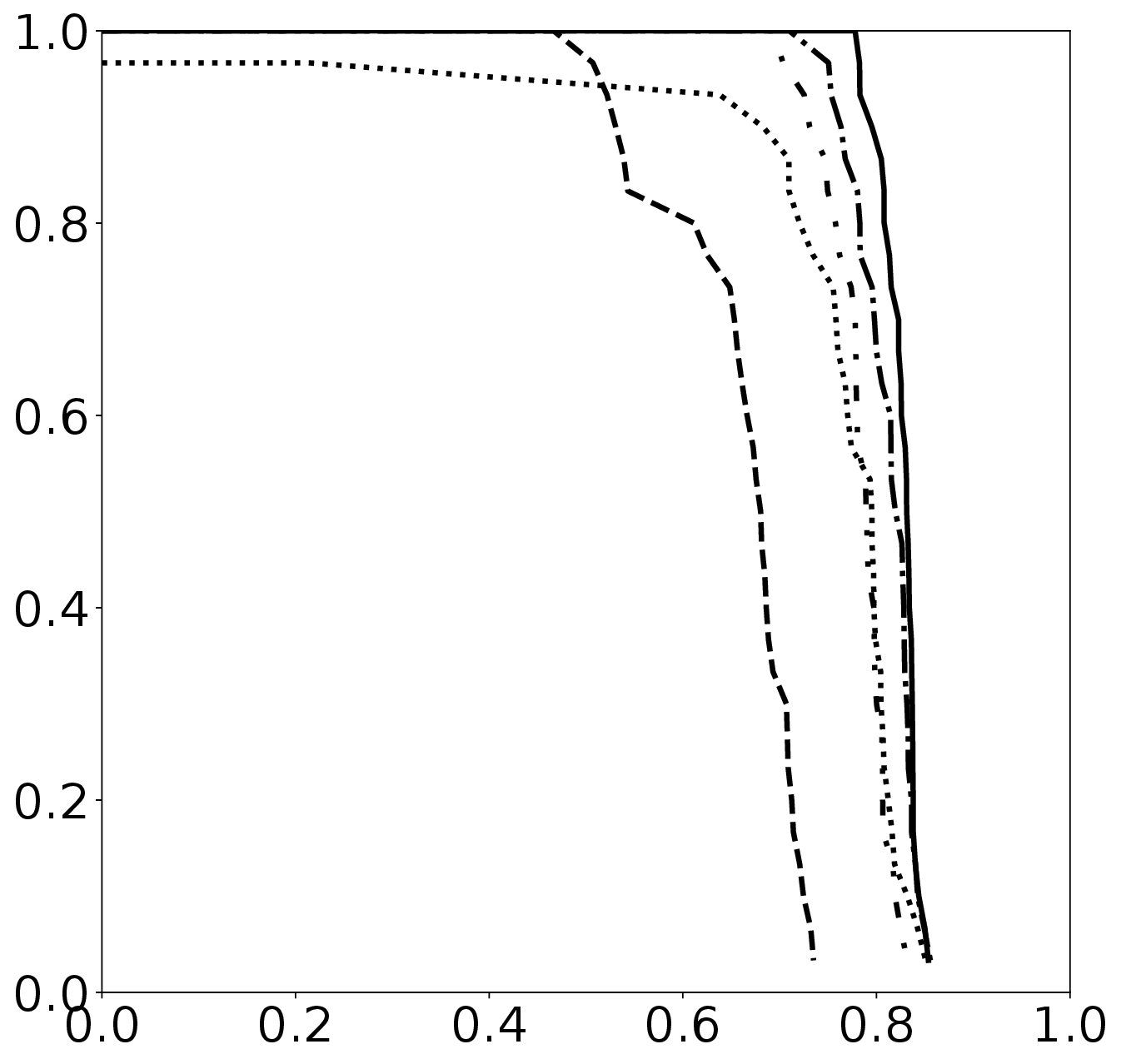}} 
\\
\subfloat[nasa 1]{\includegraphics[width=0.2\linewidth]{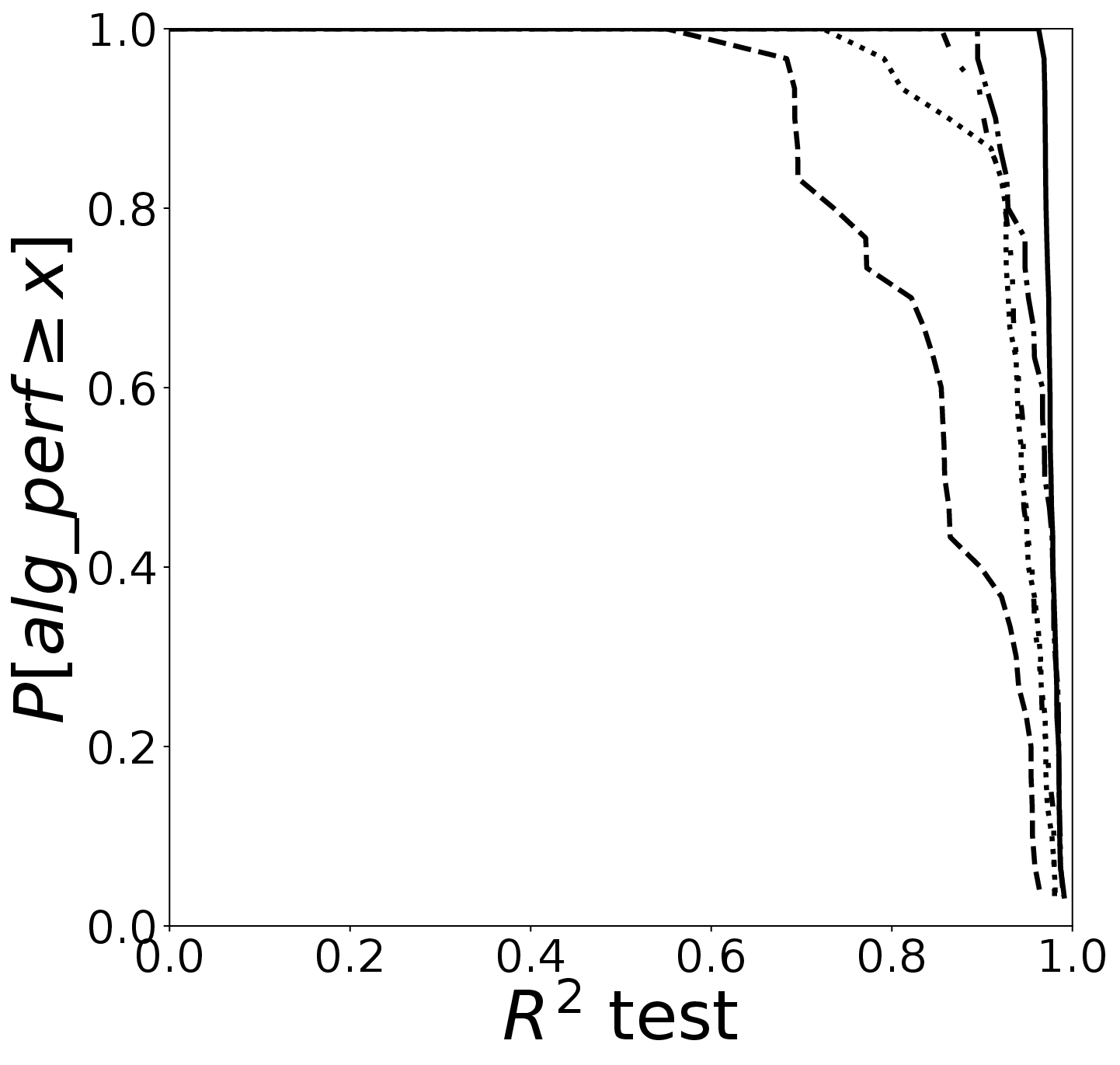}} 
\subfloat[nasa 2]{\includegraphics[width=0.19\linewidth]{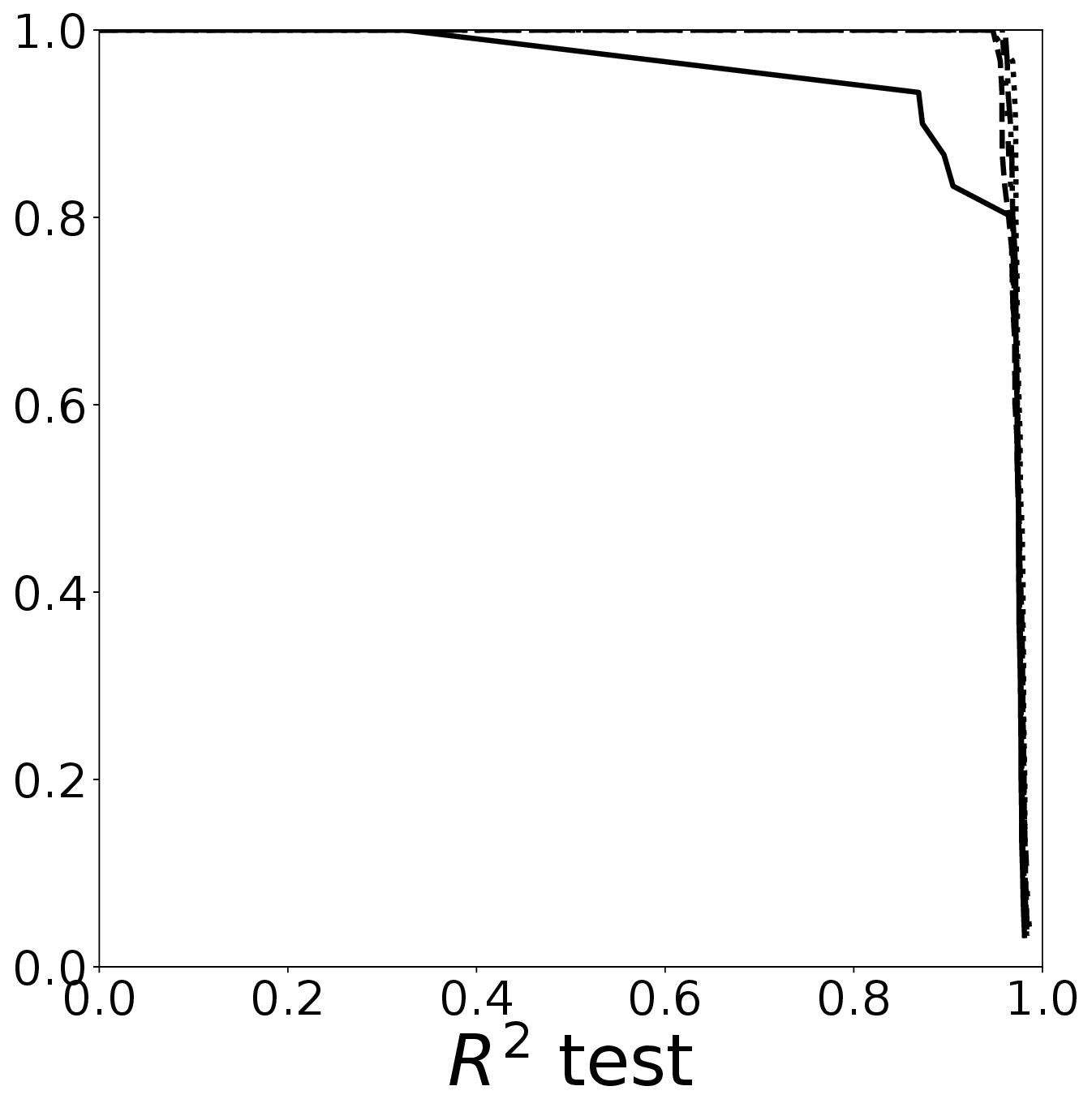}} 
\subfloat[niku 1]{\includegraphics[width=0.19\linewidth]{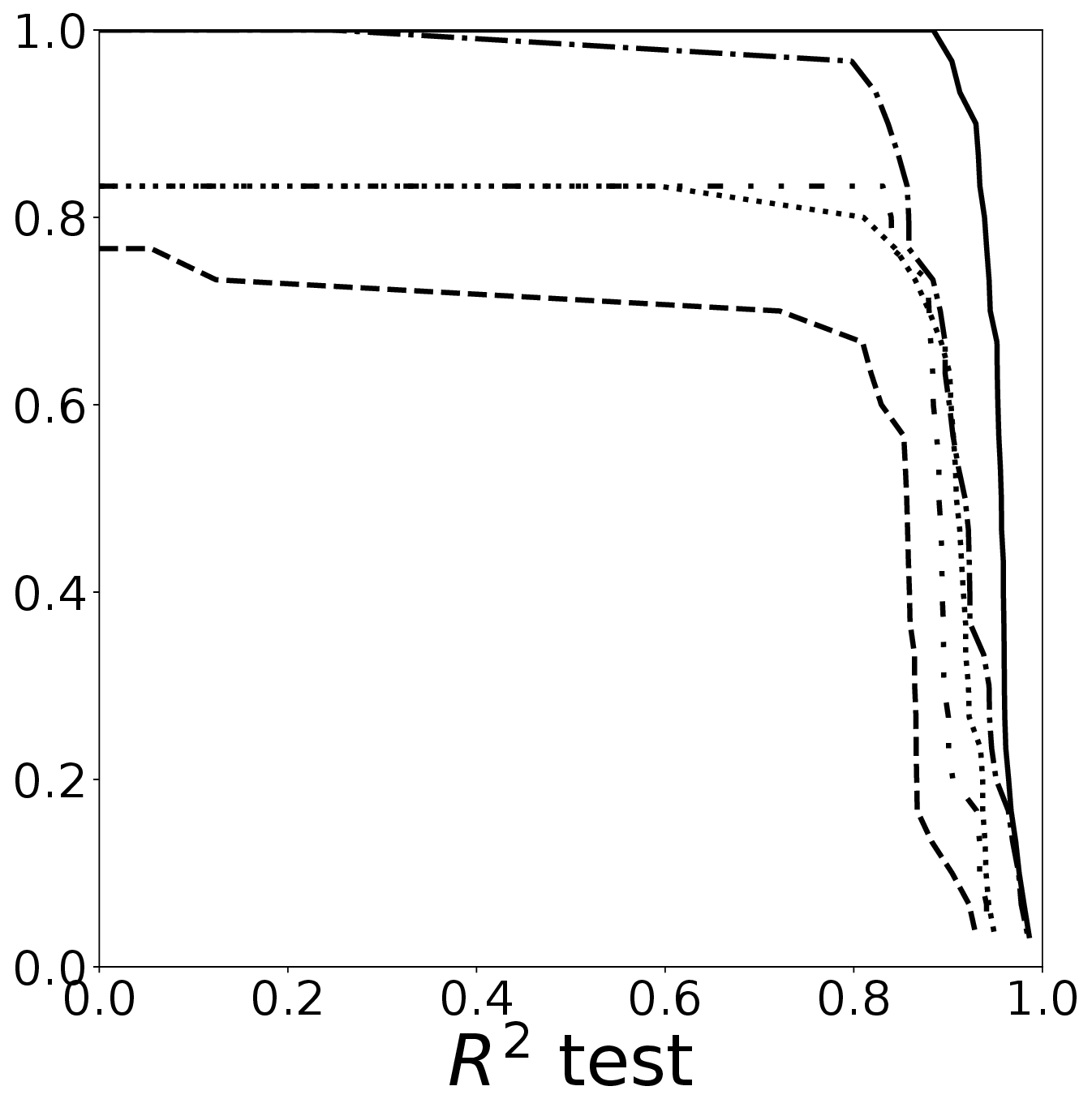}}
\subfloat[niku 2]{\includegraphics[width=0.19\linewidth]{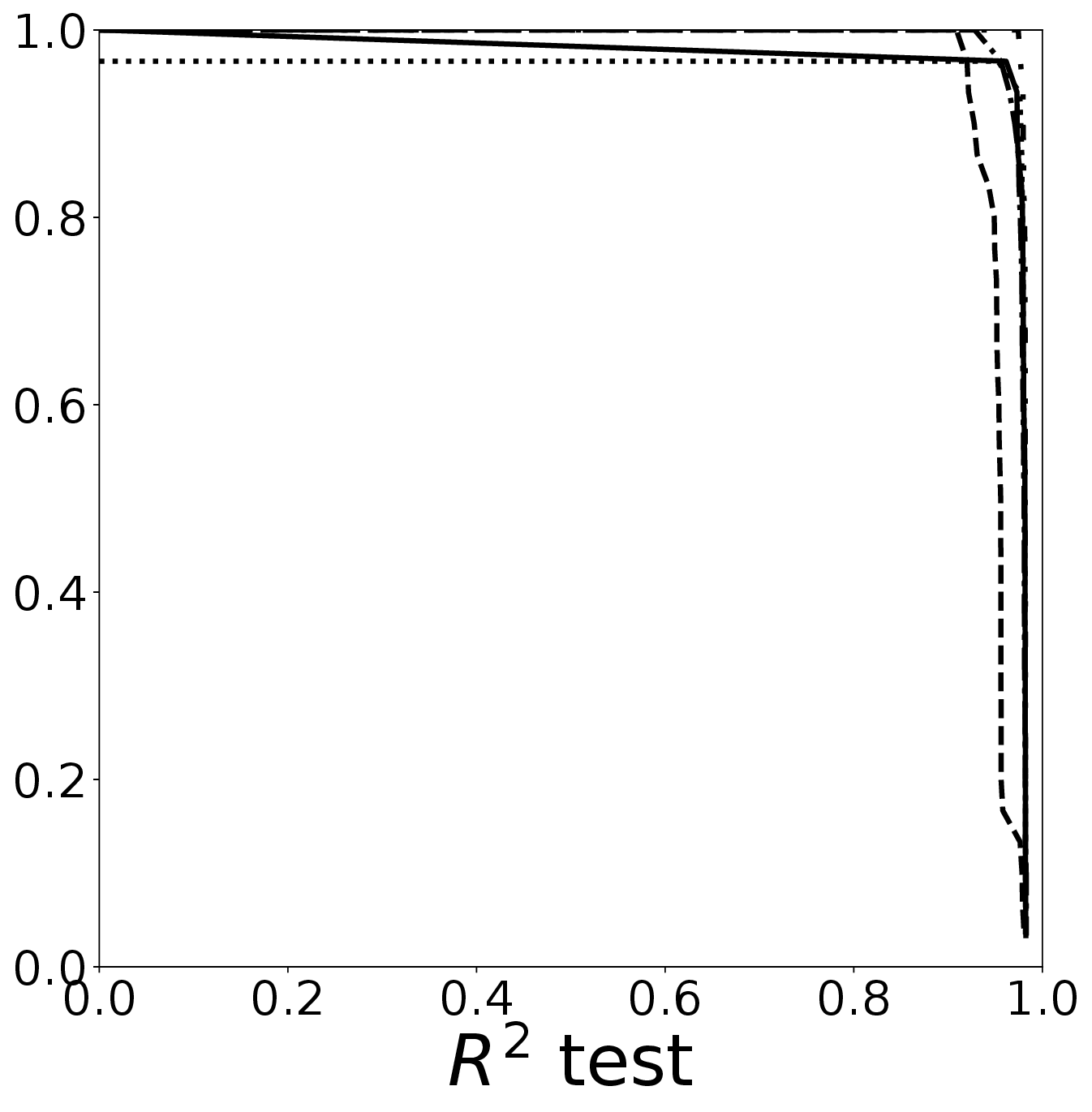}} 
\caption{Performance plots for the real-world datasets. This plot shows the probability of returning an $R^2$ equal or larger than $x$ on a random run of each algorithm.}
\label{fig:realworld_results}
\end{figure*}

In Fig.~\ref{fig:realworld_results} we observe a similar behavior for the \textbf{real-world datasets} with \egp{} covering an area close to or better than the best competing algorithm. The largest difference was on the \emph{friction} dataset in which \egp{}\textsubscript{mo} covered an area $13\%$ smaller th Operon. Considering AUC, \egp{}\textsubscript{mo}, Operon, \egp{}\textsubscript{so} obtained a similar average score and were ranked in this order. PySR and tinyGP obtained significatively worse average scores.
Considering the ranks on the median of the $R^2$ (Table~\ref{tab:rank_realworld}), the multi-objective version consistently achieved second place, but in this set, Operon was ranked first for most of the datasets. We should notice, though, that \egp{} results were always close to the best competing algorithm, while even Operon misbehaved in two datasets (flow and niku-2). When observing the statistical test results, we can conclude that there are no significant differences between the \egp{}\textsubscript{mo} and Operon, but the hypotheses of \egp{} being equivalent to PySR and tinyGP can be rejected. Regarding the model size, for this set of benchmarks in which we used a smaller maximum size, all algorithms returned a model size close to the maximum, on average. In this criteria, PySR consistently returned smaller models but with the expense of smaller accuracy as seen in the AUC results.

\begin{table}[t!]
    \centering
    \caption{Ranks of the median (1st block), AUC (2nd block), and average size (3rd block) of the test set $R^2$ for the real-world. The $p$-values were calculated with a Wilcoxon signed-rank test using as alternative hypotheses ($\alpha = 0.05$) being greater ($>$) than \egp{}.}\label{tab:rank_realworld}
    \begin{tabular}{lRRRRR}
\toprule
dataset & \text{eggp}\textsubscript{mo} & \text{eggp}\textsubscript{so} & \text{Operon} & \text{PySR} & \text{tinyGP} \\
\midrule
chemical 1 & 2 & 3 & \mathbf{1} & 5 & 4 \\
chemical 2 & 2 & 3 & \mathbf{1} & 5 & 4 \\
flow & \mathbf{1} & 2 & 5 & 3 & 4 \\
friction & 2 & 4 & \mathbf{1} & 5 & 3 \\
friction dyn & 2 & 4 & \mathbf{1} & 5 & 3 \\
nasa 1 & 2 & 3 & \mathbf{1} & 5 & 4 \\
nasa 2 & 3 & 2 & 4 & 5 & \mathbf{1} \\
niku 1 & 2 & 4 & \mathbf{1} & 5 & 3 \\
niku 2 & 4 & \mathbf{1} & 2 & 5 & 3 \\
\midrule
mean & 2.22 & 2.89 & \mathbf{1.89} & 4.78 & 3.22 \\
$p$-value $>$ & & $0.02$ & $0.93$ & $0.00$ & $0.01$ \\
\midrule \midrule
chemical 1 & 0.85 & 0.84 & \mathbf{0.89} & 0.67 & 0.81 \\
chemical 2 & 0.56 & 0.55 & \mathbf{0.67} & 0.07 & 0.37 \\
flow & \mathbf{0.99} & 0.99 & 0.57 & 0.98 & 0.88 \\
friction & 0.74 & 0.66 & \mathbf{0.83} & 0.30 & 0.65 \\
friction dyn & 0.81 & 0.78 & \mathbf{0.83} & 0.66 & 0.74 \\
nasa 1 & 0.96 & 0.94 & \mathbf{0.98} & 0.85 & 0.93 \\
nasa 2 & 0.97 & 0.97 & 0.94 & 0.97 & \mathbf{0.98} \\
niku 1 & 0.87 & 0.75 & \mathbf{0.95} & 0.62 & 0.75 \\
niku 2 & 0.98 & \mathbf{0.98} & 0.96 & 0.95 & 0.95 \\
\midrule
mean & \mathbf{0.86} & 0.83 & 0.85 & 0.67 & 0.79 \\
$p$-value $>$ & & $0.00$ & $0.85$ & $0.00$ & $0.00$ \\
\midrule 
chemical 1 & \mathbf{27.10} & 27.50 & 28.03 & 27.57 & 29.23 \\
chemical 2 & 27.40 & 26.07 & 29.07 & \mathbf{22.40} & 29.10 \\
flow & 18.80 & 23.73 & \mathbf{15.77} & 17.33 & 19.14 \\
friction & 29.83 & 26.90 & 29.23 & \mathbf{22.53} & 29.07 \\
friction dyn & 29.47 & 28.57 & 28.90 & \mathbf{24.50} & 28.97 \\
nasa 1 & 18.67 & 19.03 & 18.20 & \mathbf{17.50} & 19.43 \\
nasa 2 & 20.13 & 18.43 & 17.50 & \mathbf{14.90} & 19.50 \\
niku 1 & 19.72 & 17.23 & 19.17 & \mathbf{16.80} & 19.48 \\
niku 2 & 18.73 & 19.43 & 19.13 & \mathbf{14.73} & 19.34 \\
mean & 23.32 & 22.99 & 22.78 & \mathbf{19.81} & 23.70 \\
\midrule
mean & 23.32 & 22.99 & 22.78 & \mathbf{19.81} & 23.70 \\

\bottomrule
\end{tabular}
\end{table}

\begin{figure}[t!]
    \centering
    \includegraphics[width=0.9\linewidth, valign=t]{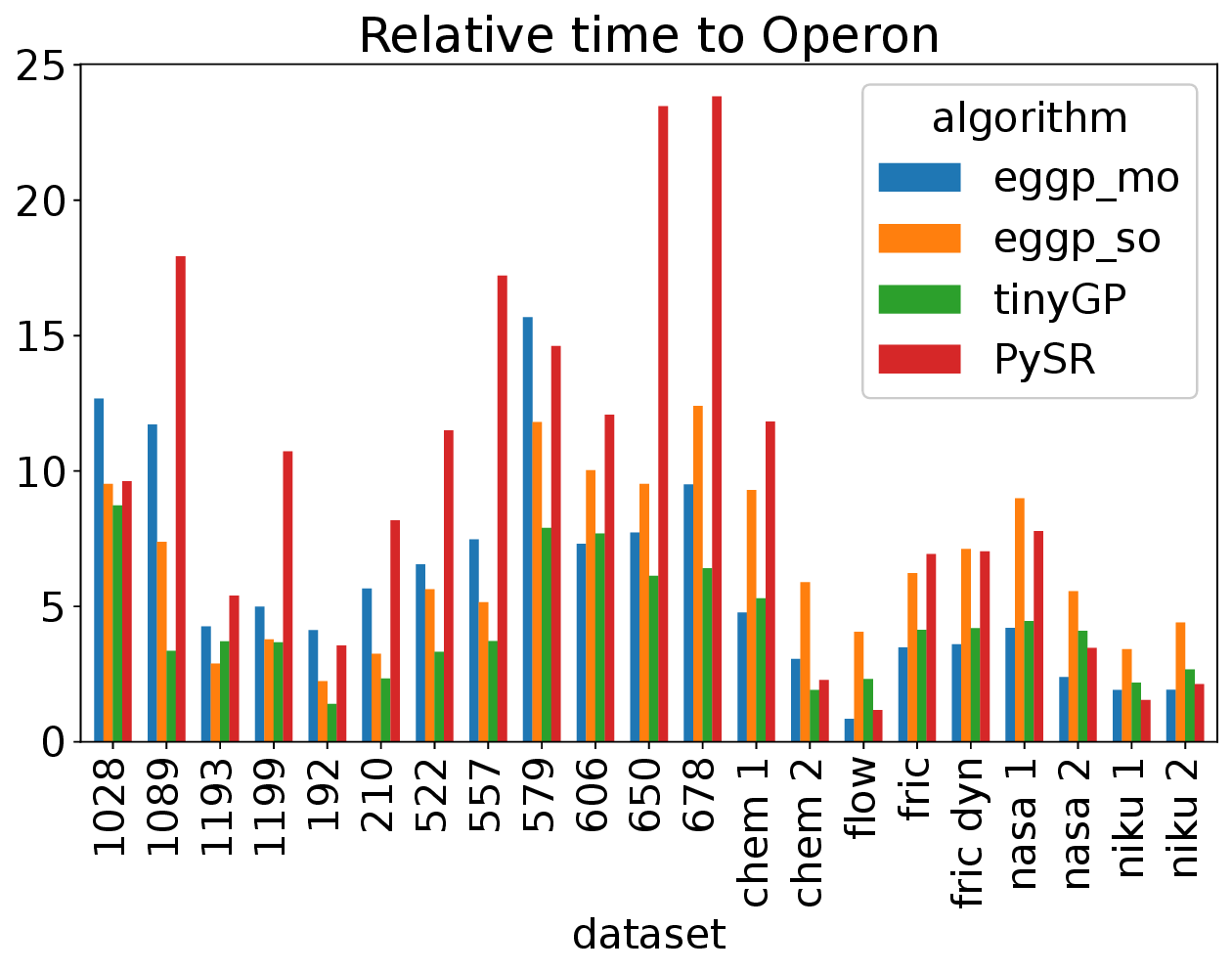}
     
    \caption{Relative avg. runtime of each algorithm using Operon as a baseline.}
    \label{fig:runtime}
\end{figure}

Regarding the \textbf{computational runtime}, since Operon is the fastest symbolic regression implementation, as noted in~\cite{burlacu2020operon}, we calculated the ratio between the average runtime of each algorithm to Operon. Fig.~\ref{fig:runtime} shows the relative runtime per dataset. In this plot we can see that \egp{} and tinyGP were both between $5$ to $15$ times slower than Operon. PySR varied from $5$ to $25$ times the runtime of Operon depending on the dataset. The higher ratios were observed on high-dimensional or larger datasets. We should stress that all algorithms were constrained to run with a single thread, thus both Operon and PySR runtime could be smaller when exploiting multi-threading.

\subsection{Additional considerations}
The main limitations of these experiments lie in the use of default or reasonable values for the hyperparameters. We should notice that \egp{} contains $8$ hyperparameters that should be fine-tuned to obtain optimal results in a practical scenario. Meanwhile, PySR contains about $30$ hyperparameters that may affect the algorithm performance\footnote{we are not considering hyperparameters that can be set using prior knowledge.} and Operon contains about $20$ hyperparameters. With careful experimentation, both PySR and Operon could have achieved similar results to those obtained with our approach.
Having said that, \egp{} is a step forward to a \emph{parameterless} experience in SR implementations, where the user only needs to set hyper-parameters that are intuitive w.r.t. their behavior. For example, increasing the number of evaluations will never make the results worse, unlike crossover and mutation rates which behaves unpredictably.

Another feature of \egp{} is the ability to export the current state of the e-graph into a file and load it when starting a new search. When loading a previous e-graph, \egp{}
will resume the search by first extracting an initial population from that e-graph. In such case, \egp{}\textsubscript{mo} will recover the current Pareto front and \egp{}\textsubscript{so} will sample random solutions. The search can be resumed with different hyper-parameters, so the user can include new non-terminals or increase the maximum size. The e-graph file can also be used with the exploration tool rEGGression\cite{rEGGression} that allows the user to explore the history of solution and retrieve a regression model outside of the Pareto front.

\subsection{Data availability}
The algorithm is implemented in Haskell using the \emph{srtree}\footnote{https://github.com/folivetti/srtree} library for symbolic regression and equality saturation.
The binaries and source code of \emph{eggp} and \emph{tinyGP} used in this paper are available at \url{https://github.com/folivetti/srtree/releases/tag/v2.0.1.0} and all the datasets, scripts, and results to replicate the experiments are available at \url{https://github.com/folivetti/eggp_paper_GECCO}.

\section{Conclusions}~\label{sec:conclusions}
In this paper we explored the use of e-graphs and equality saturation as a mechanism to keep track of the history of the search engine for symbolic regression and to exploit its pattern matching capabilities to propose a variation to the traditional subtree crossover and mutation that increases the probability of generating novel expressions.

The e-graph data structure compactly stores shared elements of a set of expressions and their equivalence relationships, and efficiently queries for parts of expressions. Exploiting this capability, we modified the subtree operators to only sample subtrees that would generate unvisited expressions. The expectation is that this simple modification would render a significant improvement in the search procedure.

We tested the proposed algorithm, called \egp{}, in $21$ different benchmarks from the literature and compared with a simple GP using the original subtree operators, and two state-of-the-art algorithms, PySR and Operon. The results showed that the modified operators are capable of improving the performance of a simple GP to compete with the state-of-the-art. The main highlight of this approach is that it consistently performs equal or better than the best competing approach.

Regarding the runtime, \egp{} is consistently faster than PySR but significantly slower than Operon. When comparing with tinyGP, we can see that the use of e-graph and equality saturation does not increase the runtime significantly. 

In conclusion, the expressiveness and capabilities of the e-graph data structure enabled us to make a simple modification to the original subtree operators while significantly improving the performance of GP for symbolic regression. This allowed us to obtain a more robust algorithm delivering better performance more reliably than state-of-the-art implementations. As for the next steps, this same modifications can be applied to any other operator used by the state-of-the-art algorithms. Not only that, but the e-graph opens up many new possibilities for improving the search as it allows us to query expressions with a combination of properties, which can translate to diversity-preserving population and easy to integrate prior knowledge~\cite{kronberger2022shape,reuter2024unit}. In addition, the storage of the search history allows us to analyze the learned building blocks and exploit this information to generate new solutions. 

\begin{acks}
F.O.F. is supported by Conselho Nacional de Desenvolvimento Cient\'{i}fico e Tecnol\'{o}gico (CNPq) grant 301596/2022-0.

G.K. is supported by the Austrian Federal Ministry for Climate Action, Environment, Energy, Mobility, Innovation and Technology, the Federal Ministry for Labour and Economy, and the regional government of Upper Austria within the COMET project ProMetHeus (904919) supported by the Austrian Research Promotion Agency (FFG). 

The authors proudly made no use of LLMs for this work.
\end{acks}

\bibliographystyle{ACM-Reference-Format}
\bibliography{references}


\begin{thebibliography}{42}


\ifx \showCODEN    \undefined \def \showCODEN     #1{\unskip}     \fi
\ifx \showDOI      \undefined \def \showDOI       #1{#1}\fi
\ifx \showISBNx    \undefined \def \showISBNx     #1{\unskip}     \fi
\ifx \showISBNxiii \undefined \def \showISBNxiii  #1{\unskip}     \fi
\ifx \showISSN     \undefined \def \showISSN      #1{\unskip}     \fi
\ifx \showLCCN     \undefined \def \showLCCN      #1{\unskip}     \fi
\ifx \shownote     \undefined \def \shownote      #1{#1}          \fi
\ifx \showarticletitle \undefined \def \showarticletitle #1{#1}   \fi
\ifx \showURL      \undefined \def \showURL       {\relax}        \fi
\providecommand\bibfield[2]{#2}
\providecommand\bibinfo[2]{#2}
\providecommand\natexlab[1]{#1}
\providecommand\showeprint[2][]{arXiv:#2}

\bibitem[Banzhaf et~al\mbox{.}(2024)]%
        {banzhaf2024combinatorics}
\bibfield{author}{\bibinfo{person}{Wolfgang Banzhaf}, \bibinfo{person}{Ting Hu}, {and} \bibinfo{person}{Gabriela Ochoa}.} \bibinfo{year}{2024}\natexlab{}.
\newblock \showarticletitle{How the Combinatorics of Neutral Spaces Leads Genetic Programming to Discover Simple Solutions}.
\newblock In \bibinfo{booktitle}{\emph{Genetic Programming Theory and Practice XX}}. \bibinfo{publisher}{Springer}, \bibinfo{pages}{65--86}.
\newblock


\bibitem[Bartlett et~al\mbox{.}(2024)]%
        {bartlett2023exhaustive}
\bibfield{author}{\bibinfo{person}{Deaglan~J. Bartlett}, \bibinfo{person}{Harry Desmond}, {and} \bibinfo{person}{Pedro~G. Ferreira}.} \bibinfo{year}{2024}\natexlab{}.
\newblock \showarticletitle{Exhaustive Symbolic Regression}.
\newblock \bibinfo{journal}{\emph{IEEE Transactions on Evolutionary Computation}} \bibinfo{volume}{28}, \bibinfo{number}{4} (\bibinfo{year}{2024}), \bibinfo{pages}{950--964}.
\newblock
\urldef\tempurl%
\url{https://doi.org/10.1109/TEVC.2023.3280250}
\showDOI{\tempurl}
\showeprint[arxiv]{2211.11461}~[astro-ph.CO]


\bibitem[Burlacu et~al\mbox{.}(2019)]%
        {burlacu2019online}
\bibfield{author}{\bibinfo{person}{Bogdan Burlacu}, \bibinfo{person}{Michael Affenzeller}, \bibinfo{person}{Gabriel Kronberger}, {and} \bibinfo{person}{Michael Kommenda}.} \bibinfo{year}{2019}\natexlab{}.
\newblock \showarticletitle{Online Diversity Control in Symbolic Regression via a Fast Hash-based Tree Similarity Measure}. In \bibinfo{booktitle}{\emph{2019 IEEE Congress on Evolutionary Computation (CEC)}}. \bibinfo{pages}{2175--2182}.
\newblock
\urldef\tempurl%
\url{https://doi.org/10.1109/CEC.2019.8790162}
\showDOI{\tempurl}


\bibitem[Burlacu et~al\mbox{.}(2020a)]%
        {burlacu2020hash}
\bibfield{author}{\bibinfo{person}{Bogdan Burlacu}, \bibinfo{person}{Lukas Kammerer}, \bibinfo{person}{Michael Affenzeller}, {and} \bibinfo{person}{Gabriel Kronberger}.} \bibinfo{year}{2020}\natexlab{a}.
\newblock \showarticletitle{Hash-Based Tree Similarity and Simplification in Genetic Programming for Symbolic Regression}. In \bibinfo{booktitle}{\emph{Computer Aided Systems Theory -- EUROCAST 2019}}, \bibfield{editor}{\bibinfo{person}{Roberto Moreno-D{\'i}az}, \bibinfo{person}{Franz Pichler}, {and} \bibinfo{person}{Alexis Quesada-Arencibia}} (Eds.). \bibinfo{publisher}{Springer International Publishing}, \bibinfo{address}{Cham}, \bibinfo{pages}{361--369}.
\newblock
\showISBNx{978-3-030-45093-9}


\bibitem[Burlacu et~al\mbox{.}(2020b)]%
        {burlacu2020operon}
\bibfield{author}{\bibinfo{person}{Bogdan Burlacu}, \bibinfo{person}{Gabriel Kronberger}, {and} \bibinfo{person}{Michael Kommenda}.} \bibinfo{year}{2020}\natexlab{b}.
\newblock \showarticletitle{Operon {C}++: an efficient genetic programming framework for symbolic regression}. In \bibinfo{booktitle}{\emph{Proceedings of the 2020 Genetic and Evolutionary Computation Conference Companion}} (Canc\'{u}n, Mexico) \emph{(\bibinfo{series}{GECCO '20})}. \bibinfo{publisher}{Association for Computing Machinery}, \bibinfo{address}{New York, NY, USA}, \bibinfo{pages}{1562–1570}.
\newblock
\showISBNx{9781450371278}
\urldef\tempurl%
\url{https://doi.org/10.1145/3377929.3398099}
\showDOI{\tempurl}


\bibitem[Cao et~al\mbox{.}(2024)]%
        {caolearning}
\bibfield{author}{\bibinfo{person}{Haotian Cao}, \bibinfo{person}{Garrett~W Merz}, \bibinfo{person}{Kyle Cranmer}, {and} \bibinfo{person}{Gary Shiu}.} \bibinfo{year}{2024}\natexlab{}.
\newblock \showarticletitle{Learning Conformal Field Theory with Symbolic Regression: Recovering the Symbolic Expressions for the Energy Spectrum}. In \bibinfo{booktitle}{\emph{NeurIPS 2024 Workshop: Machine Learning and the Physical Sciences}}.
\newblock


\bibitem[Cao et~al\mbox{.}(2023)]%
        {cao2023genetic}
\bibfield{author}{\bibinfo{person}{Lulu Cao}, \bibinfo{person}{Zimo Zheng}, \bibinfo{person}{Chenwen Ding}, \bibinfo{person}{Jinkai Cai}, {and} \bibinfo{person}{Min Jiang}.} \bibinfo{year}{2023}\natexlab{}.
\newblock \showarticletitle{Genetic Programming Symbolic Regression with Simplification-Pruning Operator for Solving Differential Equations}. In \bibinfo{booktitle}{\emph{International Conference on Neural Information Processing}}. Springer, \bibinfo{pages}{287--298}.
\newblock


\bibitem[Cranmer(2023)]%
        {cranmerpysr}
\bibfield{author}{\bibinfo{person}{Miles Cranmer}.} \bibinfo{year}{2023}\natexlab{}.
\newblock \bibinfo{title}{Interpretable Machine Learning for Science with PySR and SymbolicRegression.jl}.
\newblock
\urldef\tempurl%
\url{https://doi.org/10.48550/ARXIV.2305.01582}
\showDOI{\tempurl}


\bibitem[de~Franca and Kronberger(2023)]%
        {de2023reducing}
\bibfield{author}{\bibinfo{person}{Fabricio~Olivetti de Franca} {and} \bibinfo{person}{Gabriel Kronberger}.} \bibinfo{year}{2023}\natexlab{}.
\newblock \showarticletitle{Reducing Overparameterization of Symbolic Regression Models with Equality Saturation}. In \bibinfo{booktitle}{\emph{Proceedings of the Genetic and Evolutionary Computation Conference}}. \bibinfo{pages}{1064--1072}.
\newblock


\bibitem[de~Franca and Kronberger(2025)]%
        {rEGGression}
\bibfield{author}{\bibinfo{person}{Fabricio~Olivetti de Franca} {and} \bibinfo{person}{Gabriel Kronberger}.} \bibinfo{year}{2025}\natexlab{}.
\newblock \showarticletitle{rEGGression: an Interactive and Agnostic Tool for the Exploration of Symbolic Regression Models}. In \bibinfo{booktitle}{\emph{Proceedings of the Genetic and Evolutionary Computation Conference}} (Malaga, Spain) \emph{(\bibinfo{series}{GECCO '25})}. \bibinfo{publisher}{Association for Computing Machinery}, \bibinfo{address}{New York, NY, USA}.
\newblock
\showISBNx{9798400714658}
\urldef\tempurl%
\url{https://doi.org/10.1145/3712256.3726385}
\showDOI{\tempurl}
\showeprint[arxiv]{2501.17859}~[cs.LG]


\bibitem[de~Franca et~al\mbox{.}(2024)]%
        {de2024srbench}
\bibfield{author}{\bibinfo{person}{F.~O. de Franca}, \bibinfo{person}{M. Virgolin}, \bibinfo{person}{M. Kommenda}, \bibinfo{person}{M.~S. Majumder}, \bibinfo{person}{M. Cranmer}, \bibinfo{person}{G. Espada}, \bibinfo{person}{L. Ingelse}, \bibinfo{person}{A. Fonseca}, \bibinfo{person}{M. Landajuela}, \bibinfo{person}{B. Petersen}, \bibinfo{person}{R. Glatt}, \bibinfo{person}{N. Mundhenk}, \bibinfo{person}{C.~S. Lee}, \bibinfo{person}{J.~D. Hochhalter}, \bibinfo{person}{D.~L. Randall}, \bibinfo{person}{P. Kamienny}, \bibinfo{person}{H. Zhang}, \bibinfo{person}{G. Dick}, \bibinfo{person}{A. Simon}, \bibinfo{person}{B. Burlacu}, \bibinfo{person}{Jaan Kasak}, \bibinfo{person}{Meera Machado}, \bibinfo{person}{Casper Wilstrup}, {and} \bibinfo{person}{W.~G.~La Cavaz}.} \bibinfo{year}{2024}\natexlab{}.
\newblock \showarticletitle{SRBench++: Principled Benchmarking of Symbolic Regression With Domain-Expert Interpretation}.
\newblock \bibinfo{journal}{\emph{IEEE Transactions on Evolutionary Computation}} (\bibinfo{year}{2024}), \bibinfo{pages}{1--1}.
\newblock
\urldef\tempurl%
\url{https://doi.org/10.1109/TEVC.2024.3423681}
\showDOI{\tempurl}


\bibitem[Deb et~al\mbox{.}(2000)]%
        {deb2000nsga2}
\bibfield{author}{\bibinfo{person}{Kalyanmoy Deb}, \bibinfo{person}{Samir Agrawal}, \bibinfo{person}{Amrit Pratap}, {and} \bibinfo{person}{T. Meyarivan}.} \bibinfo{year}{2000}\natexlab{}.
\newblock \showarticletitle{A Fast Elitist Non-dominated Sorting Genetic Algorithm for Multi-objective Optimization: NSGA-II}. In \bibinfo{booktitle}{\emph{Parallel Problem Solving from Nature PPSN VI}}, \bibfield{editor}{\bibinfo{person}{Marc Schoenauer}, \bibinfo{person}{Kalyanmoy Deb}, \bibinfo{person}{G{\"u}nther Rudolph}, \bibinfo{person}{Xin Yao}, \bibinfo{person}{Evelyne Lutton}, \bibinfo{person}{Juan~Julian Merelo}, {and} \bibinfo{person}{Hans-Paul Schwefel}} (Eds.). \bibinfo{publisher}{Springer Berlin Heidelberg}, \bibinfo{address}{Berlin, Heidelberg}, \bibinfo{pages}{849--858}.
\newblock
\showISBNx{978-3-540-45356-7}


\bibitem[Ebner et~al\mbox{.}(2001)]%
        {10.1002/cplx.10021}
\bibfield{author}{\bibinfo{person}{Marc Ebner}, \bibinfo{person}{Mark Shackleton}, {and} \bibinfo{person}{Rob Shipman}.} \bibinfo{year}{2001}\natexlab{}.
\newblock \showarticletitle{How neutral networks influence evolvability}.
\newblock \bibinfo{journal}{\emph{Complexity}} \bibinfo{volume}{7}, \bibinfo{number}{2} (\bibinfo{year}{2001}), \bibinfo{pages}{19--33}.
\newblock
\urldef\tempurl%
\url{https://doi.org/10.1002/cplx.10021}
\showDOI{\tempurl}
\showeprint{https://onlinelibrary.wiley.com/doi/pdf/10.1002/cplx.10021}


\bibitem[Gustafson et~al\mbox{.}(2005)]%
        {gustafson2005improving}
\bibfield{author}{\bibinfo{person}{S. Gustafson}, \bibinfo{person}{E.K. Burke}, {and} \bibinfo{person}{N. Krasnogor}.} \bibinfo{year}{2005}\natexlab{}.
\newblock \showarticletitle{On improving genetic programming for symbolic regression}. In \bibinfo{booktitle}{\emph{2005 IEEE Congress on Evolutionary Computation}}, Vol.~\bibinfo{volume}{1}. \bibinfo{pages}{912--919 Vol.1}.
\newblock
\urldef\tempurl%
\url{https://doi.org/10.1109/CEC.2005.1554780}
\showDOI{\tempurl}


\bibitem[Hu and Banzhaf(2018)]%
        {hu2018neutrality}
\bibfield{author}{\bibinfo{person}{Ting Hu} {and} \bibinfo{person}{Wolfgang Banzhaf}.} \bibinfo{year}{2018}\natexlab{}.
\newblock \bibinfo{booktitle}{\emph{Neutrality, Robustness, and Evolvability in Genetic Programming}}.
\newblock \bibinfo{publisher}{Springer International Publishing}, \bibinfo{address}{Cham}, \bibinfo{pages}{101--117}.
\newblock
\showISBNx{978-3-319-97088-2}
\urldef\tempurl%
\url{https://doi.org/10.1007/978-3-319-97088-2_7}
\showDOI{\tempurl}


\bibitem[Hu et~al\mbox{.}(2023)]%
        {hu2023phenotype}
\bibfield{author}{\bibinfo{person}{Ting Hu}, \bibinfo{person}{Gabriela Ochoa}, {and} \bibinfo{person}{Wolfgang Banzhaf}.} \bibinfo{year}{2023}\natexlab{}.
\newblock \showarticletitle{Phenotype Search Trajectory Networks for Linear Genetic Programming}. In \bibinfo{booktitle}{\emph{Genetic Programming}}, \bibfield{editor}{\bibinfo{person}{Gisele Pappa}, \bibinfo{person}{Mario Giacobini}, {and} \bibinfo{person}{Zdenek Vasicek}} (Eds.). \bibinfo{publisher}{Springer Nature Switzerland}, \bibinfo{address}{Cham}, \bibinfo{pages}{52--67}.
\newblock
\showISBNx{978-3-031-29573-7}


\bibitem[Imai~Aldeia et~al\mbox{.}(2024)]%
        {seidyo2024inexact}
\bibfield{author}{\bibinfo{person}{Guilherme~Seidyo Imai~Aldeia}, \bibinfo{person}{Fabr\'{\i}cio~Olivetti De~Fran\c{c}a}, {and} \bibinfo{person}{William~G. La~Cava}.} \bibinfo{year}{2024}\natexlab{}.
\newblock \showarticletitle{Inexact Simplification of Symbolic Regression Expressions with Locality-sensitive Hashing}. In \bibinfo{booktitle}{\emph{Proceedings of the Genetic and Evolutionary Computation Conference}} (Melbourne, VIC, Australia) \emph{(\bibinfo{series}{GECCO '24})}. \bibinfo{publisher}{Association for Computing Machinery}, \bibinfo{address}{New York, NY, USA}, \bibinfo{pages}{896–904}.
\newblock
\showISBNx{9798400704949}
\urldef\tempurl%
\url{https://doi.org/10.1145/3638529.3654147}
\showDOI{\tempurl}


\bibitem[Keller and Banzhaf(1999)]%
        {keller1999evolution}
\bibfield{author}{\bibinfo{person}{Robert~E. Keller} {and} \bibinfo{person}{Wolfgang Banzhaf}.} \bibinfo{year}{1999}\natexlab{}.
\newblock \showarticletitle{The evolution of genetic code in Genetic Programming}. In \bibinfo{booktitle}{\emph{Proceedings of the 1st Annual Conference on Genetic and Evolutionary Computation - Volume 2}} (Orlando, Florida) \emph{(\bibinfo{series}{GECCO'99})}. \bibinfo{publisher}{Morgan Kaufmann Publishers Inc.}, \bibinfo{address}{San Francisco, CA, USA}, \bibinfo{pages}{1077–1082}.
\newblock
\showISBNx{1558606114}


\bibitem[Koza(1992)]%
        {Koza1992}
\bibfield{author}{\bibinfo{person}{John~R. Koza}.} \bibinfo{year}{1992}\natexlab{}.
\newblock \bibinfo{booktitle}{\emph{Genetic Programming: On the Programming of Computers by Means of Natural Selection}}.
\newblock \bibinfo{publisher}{MIT Press}, \bibinfo{address}{Cambridge, MA, USA}.
\newblock
\showISBNx{0-262-11170-5}


\bibitem[Kronberger et~al\mbox{.}(2024a)]%
        {kronberger2024}
\bibfield{author}{\bibinfo{person}{Gabriel Kronberger}, \bibinfo{person}{Bogdan Burlacu}, \bibinfo{person}{Michael Kommenda}, \bibinfo{person}{Stephan~M. Winkler}, {and} \bibinfo{person}{Michael Affenzeller}.} \bibinfo{year}{2024}\natexlab{a}.
\newblock \bibinfo{booktitle}{\emph{Symbolic Regression}}.
\newblock \bibinfo{publisher}{Chapman \& Hall / CRC Press}.
\newblock


\bibitem[Kronberger et~al\mbox{.}(2022)]%
        {kronberger2022shape}
\bibfield{author}{\bibinfo{person}{G. Kronberger}, \bibinfo{person}{F.~O. de Franca}, \bibinfo{person}{B. Burlacu}, \bibinfo{person}{C. Haider}, {and} \bibinfo{person}{M. Kommenda}.} \bibinfo{year}{2022}\natexlab{}.
\newblock \showarticletitle{Shape-Constrained Symbolic Regression—Improving Extrapolation with Prior Knowledge}.
\newblock \bibinfo{journal}{\emph{Evolutionary Computation}} \bibinfo{volume}{30}, \bibinfo{number}{1} (\bibinfo{date}{03} \bibinfo{year}{2022}), \bibinfo{pages}{75--98}.
\newblock
\showISSN{1063-6560}
\urldef\tempurl%
\url{https://doi.org/10.1162/evco_a_00294}
\showDOI{\tempurl}


\bibitem[Kronberger et~al\mbox{.}(2024b)]%
        {kronberger2024inefficiency}
\bibfield{author}{\bibinfo{person}{Gabriel Kronberger}, \bibinfo{person}{Fabricio Olivetti~de Franca}, \bibinfo{person}{Harry Desmond}, \bibinfo{person}{Deaglan~J. Bartlett}, {and} \bibinfo{person}{Lukas Kammerer}.} \bibinfo{year}{2024}\natexlab{b}.
\newblock \showarticletitle{The Inefficiency of Genetic Programming for Symbolic Regression}. In \bibinfo{booktitle}{\emph{Parallel Problem Solving from Nature -- PPSN XVIII}}, \bibfield{editor}{\bibinfo{person}{Michael Affenzeller}, \bibinfo{person}{Stephan~M. Winkler}, \bibinfo{person}{Anna~V. Kononova}, \bibinfo{person}{Heike Trautmann}, \bibinfo{person}{Tea Tu{\v{s}}ar}, \bibinfo{person}{Penousal Machado}, {and} \bibinfo{person}{Thomas B{\"a}ck}} (Eds.). \bibinfo{publisher}{Springer Nature Switzerland}, \bibinfo{address}{Cham}, \bibinfo{pages}{273--289}.
\newblock
\showISBNx{978-3-031-70055-2}


\bibitem[Kronberger and {Olivetti de França}(2025)]%
        {kronberger2024jsc}
\bibfield{author}{\bibinfo{person}{Gabriel Kronberger} {and} \bibinfo{person}{Fabrício {Olivetti de França}}.} \bibinfo{year}{2025}\natexlab{}.
\newblock \showarticletitle{Effects of reducing redundant parameters in parameter optimization for symbolic regression using genetic programming}.
\newblock \bibinfo{journal}{\emph{Journal of Symbolic Computation}}  \bibinfo{volume}{129} (\bibinfo{year}{2025}), \bibinfo{pages}{102413}.
\newblock
\showISSN{0747-7171}
\urldef\tempurl%
\url{https://doi.org/10.1016/j.jsc.2024.102413}
\showDOI{\tempurl}


\bibitem[La~Cava et~al\mbox{.}(2021)]%
        {la2021contemporary}
\bibfield{author}{\bibinfo{person}{William La~Cava}, \bibinfo{person}{Bogdan Burlacu}, \bibinfo{person}{Marco Virgolin}, \bibinfo{person}{Michael Kommenda}, \bibinfo{person}{Patryk Orzechowski}, \bibinfo{person}{Fabr{\'\i}cio~Olivetti de Fran{\c{c}}a}, \bibinfo{person}{Ying Jin}, {and} \bibinfo{person}{Jason~H Moore}.} \bibinfo{year}{2021}\natexlab{}.
\newblock \showarticletitle{Contemporary symbolic regression methods and their relative performance}.
\newblock \bibinfo{journal}{\emph{Advances in neural information processing systems}} \bibinfo{volume}{2021}, \bibinfo{number}{DB1} (\bibinfo{year}{2021}), \bibinfo{pages}{1}.
\newblock


\bibitem[{Lelli} et~al\mbox{.}(2017)]%
        {RAR}
\bibfield{author}{\bibinfo{person}{Federico {Lelli}}, \bibinfo{person}{Stacy~S. {McGaugh}}, \bibinfo{person}{James~M. {Schombert}}, {and} \bibinfo{person}{Marcel~S. {Pawlowski}}.} \bibinfo{year}{2017}\natexlab{}.
\newblock \showarticletitle{{One Law to Rule Them All: The Radial Acceleration Relation of Galaxies}}.
\newblock \bibinfo{journal}{\emph{Astrophysical Journal}} \bibinfo{volume}{836}, \bibinfo{number}{2}, Article \bibinfo{articleno}{152} (\bibinfo{date}{Feb.} \bibinfo{year}{2017}), \bibinfo{numpages}{152}~pages.
\newblock
\urldef\tempurl%
\url{https://doi.org/10.3847/1538-4357/836/2/152}
\showDOI{\tempurl}
\showeprint[arxiv]{1610.08981}~[astro-ph.GA]


\bibitem[McPhee et~al\mbox{.}(2008)]%
        {mcphee2008semantic}
\bibfield{author}{\bibinfo{person}{Nicholas~Freitag McPhee}, \bibinfo{person}{Brian Ohs}, {and} \bibinfo{person}{Tyler Hutchison}.} \bibinfo{year}{2008}\natexlab{}.
\newblock \showarticletitle{Semantic building blocks in genetic programming}. In \bibinfo{booktitle}{\emph{Genetic Programming: 11th European Conference, EuroGP 2008, Naples, Italy, March 26-28, 2008. Proceedings 11}}. Springer, \bibinfo{pages}{134--145}.
\newblock


\bibitem[Miller and Smith(2006)]%
        {milleratall2006}
\bibfield{author}{\bibinfo{person}{J.F. Miller} {and} \bibinfo{person}{S.L. Smith}.} \bibinfo{year}{2006}\natexlab{}.
\newblock \showarticletitle{Redundancy and computational efficiency in Cartesian genetic programming}.
\newblock \bibinfo{journal}{\emph{IEEE Transactions on Evolutionary Computation}} \bibinfo{volume}{10}, \bibinfo{number}{2} (\bibinfo{year}{2006}), \bibinfo{pages}{167--174}.
\newblock
\urldef\tempurl%
\url{https://doi.org/10.1109/TEVC.2006.871253}
\showDOI{\tempurl}


\bibitem[Moraglio et~al\mbox{.}(2012)]%
        {moraglio2012geometric}
\bibfield{author}{\bibinfo{person}{Alberto Moraglio}, \bibinfo{person}{Krzysztof Krawiec}, {and} \bibinfo{person}{Colin~G Johnson}.} \bibinfo{year}{2012}\natexlab{}.
\newblock \showarticletitle{Geometric semantic genetic programming}. In \bibinfo{booktitle}{\emph{International Conference on Parallel Problem Solving from Nature}}. Springer, \bibinfo{pages}{21--31}.
\newblock


\bibitem[Murty and Kabadi(1985)]%
        {murty1985some}
\bibfield{author}{\bibinfo{person}{Katta~G Murty} {and} \bibinfo{person}{Santosh~N Kabadi}.} \bibinfo{year}{1985}\natexlab{}.
\newblock \bibinfo{booktitle}{\emph{Some NP-complete problems in quadratic and nonlinear programming}}.
\newblock \bibinfo{type}{{T}echnical {R}eport}.
\newblock


\bibitem[Randall et~al\mbox{.}(2022)]%
        {randall2022bingo}
\bibfield{author}{\bibinfo{person}{David~L Randall}, \bibinfo{person}{Tyler~S Townsend}, \bibinfo{person}{Jacob~D Hochhalter}, {and} \bibinfo{person}{Geoffrey~F Bomarito}.} \bibinfo{year}{2022}\natexlab{}.
\newblock \showarticletitle{Bingo: a customizable framework for symbolic regression with genetic programming}. In \bibinfo{booktitle}{\emph{Proceedings of the Genetic and Evolutionary Computation Conference Companion}}. \bibinfo{pages}{2282--2288}.
\newblock


\bibitem[Reuter et~al\mbox{.}(2024)]%
        {reuter2024unit}
\bibfield{author}{\bibinfo{person}{Julia Reuter}, \bibinfo{person}{Viktor Martinek}, \bibinfo{person}{Roland Herzog}, {and} \bibinfo{person}{Sanaz Mostaghim}.} \bibinfo{year}{2024}\natexlab{}.
\newblock \showarticletitle{Unit-Aware Genetic Programming for the Development of Empirical Equations}. In \bibinfo{booktitle}{\emph{International Conference on Parallel Problem Solving from Nature}}. Springer, \bibinfo{pages}{168--183}.
\newblock


\bibitem[Rivero et~al\mbox{.}(2022)]%
        {rivero2022dome}
\bibfield{author}{\bibinfo{person}{Daniel Rivero}, \bibinfo{person}{Enrique Fernandez-Blanco}, {and} \bibinfo{person}{Alejandro Pazos}.} \bibinfo{year}{2022}\natexlab{}.
\newblock \showarticletitle{Do{M}E: A deterministic technique for equation development and Symbolic Regression}.
\newblock \bibinfo{journal}{\emph{Expert Systems with Applications}}  \bibinfo{volume}{198} (\bibinfo{year}{2022}), \bibinfo{pages}{116712}.
\newblock


\bibitem[Ruberto et~al\mbox{.}(2019)]%
        {ruberto2019genetic}
\bibfield{author}{\bibinfo{person}{Stefano Ruberto}, \bibinfo{person}{Leonardo Vanneschi}, {and} \bibinfo{person}{Mauro Castelli}.} \bibinfo{year}{2019}\natexlab{}.
\newblock \showarticletitle{Genetic programming with semantic equivalence classes}.
\newblock \bibinfo{journal}{\emph{Swarm and evolutionary computation}}  \bibinfo{volume}{44} (\bibinfo{year}{2019}), \bibinfo{pages}{453--469}.
\newblock


\bibitem[Russeil et~al\mbox{.}(2024)]%
        {russeil2024multiview}
\bibfield{author}{\bibinfo{person}{Etienne Russeil}, \bibinfo{person}{Fabr{\'\i}cio~Olivetti de Fran{\c{c}}a}, \bibinfo{person}{Konstantin Malanchev}, \bibinfo{person}{Bogdan Burlacu}, \bibinfo{person}{Emille Ishida}, \bibinfo{person}{Marion Leroux}, \bibinfo{person}{Cl{\'e}ment Michelin}, \bibinfo{person}{Guillaume Moinard}, {and} \bibinfo{person}{Emmanuel Gangler}.} \bibinfo{year}{2024}\natexlab{}.
\newblock \showarticletitle{Multiview Symbolic Regression}. In \bibinfo{booktitle}{\emph{Proceedings of the Genetic and Evolutionary Computation Conference}}. \bibinfo{pages}{961--970}.
\newblock


\bibitem[Schmidt and Lipson(2009)]%
        {schmidt2009distilling}
\bibfield{author}{\bibinfo{person}{Michael Schmidt} {and} \bibinfo{person}{Hod Lipson}.} \bibinfo{year}{2009}\natexlab{}.
\newblock \showarticletitle{Distilling free-form natural laws from experimental data}.
\newblock \bibinfo{journal}{\emph{science}} \bibinfo{volume}{324}, \bibinfo{number}{5923} (\bibinfo{year}{2009}), \bibinfo{pages}{81--85}.
\newblock


\bibitem[Sipper(2019)]%
        {Sipper2019tinyGP}
\bibfield{author}{\bibinfo{person}{M. Sipper}.} \bibinfo{year}{2019}\natexlab{}.
\newblock \bibinfo{title}{Tiny Genetic Programming in {P}ython}.
\newblock \bibinfo{howpublished}{\url{https://github.com/moshesipper/tiny_gp}}.
\newblock


\bibitem[Tate et~al\mbox{.}(2009)]%
        {tate2009equality}
\bibfield{author}{\bibinfo{person}{Ross Tate}, \bibinfo{person}{Michael Stepp}, \bibinfo{person}{Zachary Tatlock}, {and} \bibinfo{person}{Sorin Lerner}.} \bibinfo{year}{2009}\natexlab{}.
\newblock \showarticletitle{Equality saturation: a new approach to optimization}. In \bibinfo{booktitle}{\emph{Proceedings of the 36th annual ACM SIGPLAN-SIGACT symposium on Principles of programming languages}}. \bibinfo{pages}{264--276}.
\newblock


\bibitem[Udrescu and Tegmark(2020)]%
        {udrescu2020ai}
\bibfield{author}{\bibinfo{person}{Silviu-Marian Udrescu} {and} \bibinfo{person}{Max Tegmark}.} \bibinfo{year}{2020}\natexlab{}.
\newblock \showarticletitle{AI Feynman: A physics-inspired method for symbolic regression}.
\newblock \bibinfo{journal}{\emph{Science Advances}} \bibinfo{volume}{6}, \bibinfo{number}{16} (\bibinfo{year}{2020}), \bibinfo{pages}{eaay2631}.
\newblock


\bibitem[Uy et~al\mbox{.}(2010)]%
        {uy2010semantic}
\bibfield{author}{\bibinfo{person}{Nguyen~Quang Uy}, \bibinfo{person}{Michael O’Neill}, \bibinfo{person}{Nguyen~Xuan Hoai}, \bibinfo{person}{Bob Mckay}, {and} \bibinfo{person}{Edgar Galv{\'a}n-L{\'o}pez}.} \bibinfo{year}{2010}\natexlab{}.
\newblock \showarticletitle{Semantic similarity based crossover in GP: The case for real-valued function regression}. In \bibinfo{booktitle}{\emph{Artifical Evolution: 9th International Conference, Evolution Artificielle, EA, 2009, Strasbourg, France, October 26-28, 2009. Revised Selected Papers 9}}. Springer, \bibinfo{pages}{170--181}.
\newblock


\bibitem[Vanneschi et~al\mbox{.}(2014)]%
        {vanneschi2014survey}
\bibfield{author}{\bibinfo{person}{Leonardo Vanneschi}, \bibinfo{person}{Mauro Castelli}, {and} \bibinfo{person}{Sara Silva}.} \bibinfo{year}{2014}\natexlab{}.
\newblock \showarticletitle{A survey of semantic methods in genetic programming}.
\newblock \bibinfo{journal}{\emph{Genetic Programming and Evolvable Machines}}  \bibinfo{volume}{15} (\bibinfo{year}{2014}), \bibinfo{pages}{195--214}.
\newblock


\bibitem[Virgolin and Pissis(2022)]%
        {virgolinsymbolic}
\bibfield{author}{\bibinfo{person}{Marco Virgolin} {and} \bibinfo{person}{Solon~P Pissis}.} \bibinfo{year}{2022}\natexlab{}.
\newblock \showarticletitle{Symbolic Regression is {NP}-hard}.
\newblock \bibinfo{journal}{\emph{Transactions on Machine Learning Research}} (\bibinfo{year}{2022}).
\newblock
\showISSN{2835-8856}
\urldef\tempurl%
\url{https://openreview.net/forum?id=LTiaPxqe2e}
\showURL{%
\tempurl}


\bibitem[Willsey et~al\mbox{.}(2021)]%
        {willsey2021egg}
\bibfield{author}{\bibinfo{person}{Max Willsey}, \bibinfo{person}{Chandrakana Nandi}, \bibinfo{person}{Yisu~Remy Wang}, \bibinfo{person}{Oliver Flatt}, \bibinfo{person}{Zachary Tatlock}, {and} \bibinfo{person}{Pavel Panchekha}.} \bibinfo{year}{2021}\natexlab{}.
\newblock \showarticletitle{Egg: Fast and extensible equality saturation}.
\newblock \bibinfo{journal}{\emph{Proceedings of the ACM on Programming Languages}} \bibinfo{volume}{5}, \bibinfo{number}{POPL} (\bibinfo{year}{2021}), \bibinfo{pages}{1--29}.
\newblock


\end{thebibliography}

\end{document}